%% file: ReviewTemplate.tex
\documentclass[10pt,twocolumn,letterpaper]{article}

\usepackage{graphicx}
\usepackage{amsmath}
\usepackage{amssymb}
\usepackage{booktabs}
\usepackage{microtype}      
\usepackage{xcolor}         
\usepackage{algorithm, algpseudocode}
\usepackage{multirow}
\usepackage{graphicx}
\usepackage{subcaption}
\usepackage{url}

\usepackage{makecell}
\usepackage[pagebackref,breaklinks,colorlinks]{hyperref}

\usepackage[capitalize]{cleveref}
\crefname{section}{Sec.}{Secs.}
\Crefname{section}{Section}{Sections}
\Crefname{table}{Table}{Tables}
\crefname{table}{Tab.}{Tabs.}

\begin{document}

\newcommand{\ourmethod}{CheckSel}
\newcommand{\ourmethodv}{Simsel}
\title{\ourmethod: Efficient and Accurate Data-valuation  \protect\\  Through Online Checkpoint Selection}
\date{}
\author{Soumi Das \\ IIT Kharagpur\\
\and
Manasvi Sagarkar \\ IIT Kharagpur \\
\and
Suparna Bhattacharya \\ Hewlett Packard Labs \\
\and
Sourangshu Bhattacharya \\ IIT Kharagpur\\
}
\maketitle

\begin{abstract}
Data valuation and subset selection have emerged as valuable tools for application-specific selection of important training data. However, the efficiency-accuracy tradeoffs of state-of-the-art methods hinder their widespread application to many AI workflows. In this paper, we propose a novel 2-phase solution to this problem. Phase 1 selects representative checkpoints from an SGD-like training algorithm, which are used in phase 2 to estimate the approximate the training data values, e.g. decrease in validation loss due to each training point. A key contribution of this paper is CheckSel, an Orthogonal Matching Pursuit-inspired online sparse approximation algorithm for checkpoint selection in the online setting, where the features are revealed one at a time. Another key contribution is the study of data valuation in the domain adaptation setting, where a data value estimator obtained using checkpoints from training trajectory in source domain training dataset is used for data valuation in a target domain training dataset. Experimental results on benchmark datasets show the proposed algorithm outperforms recent baseline methods by upto $\sim 30\%$ in terms of test accuracy while incurring similar computational burden, for both standalone and domain adaptation settings.

\end{abstract}

\input{intro}
\input{related}
\input{method3}

\input{experiment1}

\input{conclude}
{\small
\bibliographystyle{ieee_fullname}
\bibliography{egbib}
}

\end{document}

%% file: intro.tex
\section{Introduction}
Finding ``influential'' datapoints in a training dataset, also known as data valuation \cite{koh2017understanding,NEURIPS2020_e6385d39} and data-subset selection \cite{10.1007/978-3-030-86520-7_41}, has emerged as an important sub-problem for many modern deep learning application domains, e.g. Data-centric AI\cite{ghorbani2019data}, Explainability in trusted AI \cite{ribeiro2016should},  \cite{arya2019explanation}, debugging the training process \cite{koh2017understanding}, scalable supervised deep learning \cite{killamsetty2021grad}. Parallely, various experiment management systems (EMS) e.g. MLFlow \cite{mlflow}, WandB \cite{wandb} are being developed to address the complexities of organisation-scale experimentation and development of AI systems. The EMS systems can store ``artifacts'' of AI workflows, e.g. small summaries of intermediate datasets, for optimization of the AI workflow \cite{derakhshan2020optimizing}. In principle, the data valuation methods could be used to select short summaries of such intermediate datasets. However, the efficiency-accuracy tradeoff of the data valuation techniques prohibit their application in such systems. In this paper, we strive to improve this tradeoff so as to make such applications practical. 

While many state-of-the art data valuation methods \cite{10.1007/978-3-030-86520-7_41,yoon2020data} are able to identify high-value (test accuracy) and small-sized subsets of training data, their computational cost is prohibitive for the current application. Recently, TracIn \cite{NEURIPS2020_e6385d39} proposed to utilize the training trajectory of SGD algorithm to estimate the reduction in validation loss (value function) due to each training datapoint. While TracIn is computationally inexpensive, the quality of data valuation depends on the training checkpoints used for approximation of value function, which is inaccurate due to multiple updates performed in between checkpoints.
In this paper, we propose to overcome these drawbacks by learning a sparse estimator for the value function based on second order approximation from selected checkpoints. We extend the orthogonal matching pursuit (OMP) \cite{4385788} technique for greedy online sparse approximation, where the features are checkpoints corresponding to the training algorithm. However, unlike in traditional sparse approximation, the features are also revealed in an online manner. This leads us to the online sparse approximation setting \cite{6319766}, \cite{saad2021online}, which is surprisingly not very well studied. One of the main contributions of this paper is to propose \textit{CheckSel}, an \textit{online sparse approximation} algorithm for checkpoint selection. Another drawback of data valuation-based subset selection algorithms is that the selected subset often lacks diversity \cite{10.1007/978-3-030-86520-7_41}. Our second contribution in this paper is the \textit{SimSel} algorithm, which given the contribution score-vectors (over validation datapoints) for different training datapoints, selects the most diverse subset of training data using a sub-modular subset selection formulation \cite{mirzasoleiman2020coresets}.

Our proposed architecture for data valuation / subset selection works in two phases: (1) \textit{checkpoint selection} from training trajectory using value function which is computationally expensive, and (2) \textit{data valuation / subset selection} using selected checkpoints, which is relatively computationally inexpensive. This segregation presents the opportunity for faster data valuation on related \textit{target} datasets, if the checkpoints have been selected on \textit{source} datasets - a setting we call \textit{domain adaptation for data valuation}. However, the checkpoint selection phase needs access to target domain data for accurate estimate of value function. In another contribution, we present the \textbf{CheckSel-Source-DA} scheme which performs checkpoint selection using the value function from source domain data, followed by a domain adaptation training using selected source as well as target domain data. This scheme can be used in EMS platforms for storing selected subsets of related datasets, thus potentially speeding up the AI-workflows with minimal decrease in accuracy.


Experimental results show that training data selected by CheckSel outperforms that by TracIn, sometimes by $\sim 30\%$ in terms of test accuracy, while consuming similar or less time. Moreover, the SimSel subset selection algorithm also substantially improves the test accuracy over vanilla TracIn scores, even when used with uniformly sampled checkpoints. 
For the domain adaptation setting, we show that the both \textbf{CheckSel-Source} (Checkpoint selection using source domain value function) and \textbf{CheckSel-Source-DA} improve over model accuracy on target domain data, over TracIn by upto $\sim 30\%$.

%% file: related.tex
\section{Related Work}
Finding influential training datapoints or data valuation has become a recent research challenge in the context of deep learning. It finds its application in subset selection, providing explanations to predictions, diagnosing mislabelled examples and so on. There have been several works in the literature of data valuation encompassing influence functions \cite{koh2017understanding}, shapley values \cite{ghorbani2019data}, reinforcement learning \cite{yoon2020data}, differentiable convex programming \cite{10.1007/978-3-030-86520-7_41}, tracking training trajectories \cite{NEURIPS2020_e6385d39} and more. However, apart from \cite{koh2017understanding} and \cite{NEURIPS2020_e6385d39}, all the other above-mentioned techniques are expensive to be scaled on large datasets and models since they merge the training and scoring datapoints in a combined framework.

Trained model information is used in \cite{koh2017understanding} and \cite{NEURIPS2020_e6385d39} to score datapoints. While \cite{koh2017understanding} estimates the change in parameter $\theta$ by upweighting the datapoint $x_i$ by a small $\epsilon$ to evaluate its influence, \cite{NEURIPS2020_e6385d39} on the other hand, tracks the training trajectory and estimates the influence of the datapoints. Our work is in close proximity to \cite{NEURIPS2020_e6385d39} with the difference being in the tracking of trajectories. While \cite{NEURIPS2020_e6385d39} uses uniformly selected checkpoints to evaluate scores, we develop an algorithm to select checkpoints during the training procedure aimed to optimise a value function.

Several works on selection of data follow two dimensions: one that learn complex selection criteria  and are jointly trained with the end-task \cite{lan2018ffnet},\cite{wu2019adaframe} and another that are designed to optimise objective functions range from using convex approach \cite{das2021tmcoss}, to submodular optimisation \cite{mirzasoleiman2020coresets}, to orthogonal matching pursuit (OMP) algorithms \cite{killamsetty2021grad}. The first class is typically expensive whereas the latter one can work in practical scenarios. Following the second class of techniques, we have developed an online orthogonal matching pursuit algorithm for the selection of checkpoints. There have been few works on online OMP algorithms \cite{saad2021online} \cite{6319766}, which mainly add instances independent of past queried instances, in an online streaming setting. Our proposed online OMP, however tracks the already selected checkpoints to make an informed decision on the incoming ones. To the best of our knowledge, this is one of the first works to develop an algorithm for selection of checkpoints while tracking the training procedure, and use them to evaluate training datapoints making the process efficient and scalable. 

%% file: method3.tex
\newcommand{\cZ}{\mathcal{Z}}
\newcommand{\cT}{\mathcal{T}}
\newcommand{\cS}{\mathcal{S}}
\newcommand{\cB}{\mathcal{B}}
\newcommand{\cA}{\mathcal{A}}
\newcommand{\cR}{\mathcal{R}}
\newcommand{\RR}{\mathbb{R}}
\newtheorem{theorem}{Theorem}
\section{Online Checkpoint Selection  for Data Valuation}
\label{sec:method}

In this section, we introduce the proposed data valuation method \textit{\ourmethod} -- a two-phase selection method using the training run of a gradient descent algorithm. The first phase selects informative checkpoints from the training algorithm, and  the second phase computes the value of each datapoint. 

\subsection{Background and Problem Definition}

Let $\mathcal{Z} = \{ z_i = (x_i , y_i)| i = 1,2,..,N\}$ be the training dataset for a supervised learning task with features $x_i \in \mathcal{X}$ and labels $y_i \in \mathcal{Y}$. Also, let $\mathcal{Z'} = \{ z'_j = (x'_j , y'_j)| j = 1,2,..,M\}$ be the validation dataset which is used for assigning values/scores to a training datapoint $z\in \mathcal{Z}$. Data valuation assigns a score $v(z)$ to a training datapoint $z$ such that $\sum_{i=1}^{N} v(z_i) = V(\mathcal{Z})$, which is the total value of the training dataset. A commonly used valuation function is the total loss on a validation dataset $\mathcal{Z'}$ \cite{ghorbani2019data}. TracIn \cite{NEURIPS2020_e6385d39} uses a 
trajectory of SGD updates to define a surrogate value function -- the decrease in loss during the training.

Let $\theta_1,...,\theta_T$ be a sequence of parameters generated by training with a given deep learning model on training dataset $\cZ$ using a mini-batch SGD-like algorithm.
Given a validation datapoint $z'$, the first order approximation for the per epoch decrease in loss, $\Delta_t(z')$ is given by:
\begin{equation}
   \Delta_t(z') = l(\theta_{t+1},z') - l(\theta_t,z') =  \nabla l(\theta_t, z').(\theta_{t+1} - \theta_t)
\label{eqn:ap1}
\end{equation}
where $l(\theta_t,z')$ is the loss(value function) on $z'$ using end-of-epoch model parameter $\theta_t$. The influence of any training point $z$ on the valuation datapoint $z'$, termed as \textit{TracIn}$(z,z')$, uses parameters ($\theta_t$) from uniformly selected end-of-epoch checkpoints $t = 1,2,..,T$ of a trained model and is defined as:
\begin{equation}
    TracIn(z,z') = \sum_{t=1}^T \eta_t \nabla_{\theta_t} l (\theta_t, z).\nabla_{\theta_t} l (\theta_t, z')
\label{eq:tracin}
\end{equation}


As shown in section \ref{sec:addgreater}, the residual error of estimates given by this method is very poor. There are two main reasons for the poor performance.
Firstly, TracIn considers only the first order approximation, while many training algorithms use second order gradient information. 
Secondly, it does not consider the intermediate checkpoints while approximating $(\theta_{t+1} - \theta_t)$ (from first order) to $\nabla_{\theta_t} l (\theta_t, z)$, that shall be necessary in a mini-batch SGD setup. Next we propose a sparse approximation based formulation to alleviate these drawbacks.


\subsection{Sparse Approximation for Checkpoint Selection}

In order to consider the effect of intermediate checkpoints, we redefine the checkpoints in a training trajectory as $\theta_t^i$, where $t=1,...,T$ are the epoch indices, and $i=1,...,O$ are the minibatch indices.
Hence, we can define the per epoch decrease in loss function as:

\begin{equation}
  \Delta_t(z') = \sum_{i=1}^O l(\theta_{t}^i,z') - l(\theta_{t}^{i-1},z') 
\label{eqn:ap2}
\end{equation}

Following Taylor series expansion, and approximating the quadratic form of Hessian matrix with the square of the first order approximation, the second order expansion becomes: $  \Delta_{t} (z') \approx -\eta \sum_{i=1}^O \nabla l(\theta_{t}^{i-1},Z_i)^T \nabla l(\theta_{t}^{i-1}, z') + \frac{1}{2}*(\nabla l(\theta_{t}^{i-1},Z_i)^T \nabla l(\theta_{t}^{i-1}, z'))^2 $. This approximation is a sum of terms, which can be thought as features calculated from the $(i,t)^{th}$ checkpoint. This leads to our definition of the features for the sparse approximation:
\begin{align}
\mathcal{C}_{i}^t(Z_{i},z') =  &  \nabla l(\theta_{t}^{i-1},Z_i)^T \nabla l(\theta_{t}^{i-1}, z') + \nonumber \\ 
 & \frac{1}{2}*(\nabla l(\theta_{t}^{i-1},Z_i)^T \nabla l(\theta_{t}^{i-1}, z'))^2 
\label{eqn:ap6}
\end{align}
Note that here, $l(\theta_{t}^{i-1},Z_i)$ is the total loss over all examples $z$ in minibatch $i$.

This definition naturally leads to our sparse approximation formulation for estimating $\Delta_t(z')$ as $\Delta_t(z') \approx \sum_{i \in S_{t}}({\alpha_i \mathcal{C}_{i}^t(Z_i,z')})$, where $S_t$ is the set of checkpoints selected from epoch $t$ (which are to be stored for future estimation) and $\alpha_i$ are the learned coefficients for each of the selected checkpoints. We denote $S = \cup_{t=1}^T S_t$ as the total set of checkpoints. Denoting $\Vec{\Delta}$ as the vector of all $\Delta_t(z')$, and $\vec{\mathcal{C}}_i(Z_i)$ analogously, we arrive at our formulation for sparse estimation as:

\begin{equation}
    \min_{S,\alpha} ||\sum_{i \in S}({\alpha_i \Vec{\mathcal{C}}_{i}^t(Z_i)}) - \Vec{\Delta}||_2 \hspace{3mm} s.t. |S| \leq k
\end{equation}
where $\alpha$ denotes the set of all $\alpha_i$. Next we propose an OMP based algorithm to solve this problem.


\subsection{Online OMP based checkpoint selection}
\label{sec:oomp}

Orthogonal matching pursuit (OMP)\cite{4385788} is a popular algorithm for solving the above mentioned sparse approximation formulation of checkpoint selection. Among its key advantages is the fact that it is a greedy method, and can be used in online mode when the datapoints $(\Delta_t(z'), \mathcal{C}_{i}^t(Z_i,z'))$ are revealed one-at-a-time. However, the current problem has an added complication: the features $\mathcal{C}_{i}^t(Z_i,z')$ are also made available in an online manner. Hence, vanilla OMP algorithm cannot be applied.

In order to motivate the proposed online OMP algorithm, we define the cumulative $\Delta_t$ vector that denotes the cumulative decrease in loss across epochs, 
\begin{equation}
I_t = \sum_{j=1}^t \Delta_j\ \ \forall t=1,...,T
\end{equation}
For notational simplicity, we drop the index corresponding to $z'$. In the online OMP setting, at the end of epoch $t$ we want to approximate:
$$ I_t \approx \sum_{j\in \cS_t} \alpha_j C_j $$ 
where, $\cS_t$ is the set of selected checkpoint indices till epoch $t$, and $C_j$ are the features for $j^{th}$ selected checkpoint.



The key ideas behind proposed online OMP algorithm (described in Algorithm \ref{algo:onlineomp}) are: (1) to keep track of the selected checkpoints over time using the variable $\cS_t$, and (2) greedily replace them with incoming checkpoints ($\mathcal{C}_i^t$) based on their projections on the residual vector ($\rho$), which is the difference between the obtained estimate ($\xi = \sum_{i\in \cS} \alpha_i C_i$) and the ground truth ($I^t$ in this case). Detailed descriptions of the overall algorithm and the replacement subroutine are provided in Algorithms \ref{algo:onlineomp} and \ref{algo:replace}, respectively. 


\newcommand{\algocomment}[1]{\textcolor{blue}{#1}}
 \begin{algorithm}[h]
 \caption{: \textbf{CheckSel}}
 \label{algo:onlineomp}
 \small
 \begin{algorithmic}[1]
\State{\textbf{Input}:} $k$: Total number of checkpoints to be selected\\
$I_t$: Cumulative decrease in loss after epoch $t$ \\
$C_i^t$: Features calculated at checkpoint $(i,t)$ (Eqn: \ref{eqn:ap6})

 \State \textbf{Initialize:} \\
  $\cS$ = $\phi$ \ \algocomment{//Set of selected checkpoints} \\
  $\alpha = zero_{|S|}$ \ \algocomment{//Set of weights for the selected checkpoints} \\
 Estimate $\xi = \sum_{i\in \{1,...,|\cS|\}} \alpha_i C_i$, initially 0
 
 \For{t = 1,...,T}
 \State \textbf{Input:} $I_t, C_i^t , \forall i\in\{1,...,O\} , ||C_i||_2 = 1 $
 \State{\textbf{Process}:}
 \State  Update $\alpha$: $\alpha = argmin_\alpha \|\sum_{i \in |S|}(\alpha_i C_i) - I^t||_2$
 \State Update $\xi$: $\xi = \sum_{i\in \cS} \alpha_i C_i$
 \For{each batch $i \in \{1,2, \dots ,O\}$:}
 \State \algocomment{/* Replacing selected checkpoints */}
 \If{$|\cS| = k$}
 \State $\cS , \alpha, \xi$ = \textbf{CheckReplace($I^t , \xi, \cS, \alpha , C_i^t$)}
 \Else
 \State \algocomment{/* Add the checkpoints sequentially */}
 \State  $\cS \longleftarrow \cS \cup \{C_i^t\}$
 \State  Update $\alpha$ ; Update $\xi$
 \EndIf
 \EndFor
 \EndFor
 \State \textbf{Output}:Final set of selected checkpoints , $\cS$
 \end{algorithmic}
\end{algorithm}
\begin{algorithm}[h]
\small
 \caption{: \textbf{CheckReplace ($I^t , \xi, \cS, \alpha , C_i^t$)}}
 \label{algo:replace}
 \begin{algorithmic}[1]
\State  $\rho = I^t - \xi$ \ \algocomment{//residual vector}
 \State $maxproj = -\infty$ \  \algocomment{// highest projected residual from $\cS$}
 \State $ind = \phi$ \  \algocomment{//index of the checkpoint to be removed  from $\cS$}
 \For{each index $j$ in $\cS$:}
 \State  $n\rho = \rho + \alpha_j*C_j$ \  \algocomment{// new residual on removing $j$}
 \State  $\hat{n\varrho} = C_i^t . n\rho$  \  \algocomment{//projection of $C_i^t$ on the new residual}
 \State  $\hat{\varrho} = C_j . n\rho$  \  \algocomment{//projection of $C_j$ on the new residual}
  \If{$\hat{n\varrho} > \hat{\varrho}$ and $\hat{n\varrho}>maxproj$}
 \State maxproj $\longleftarrow$  $\hat{\varrho}$
 \State $ind = j$
 \EndIf
 \EndFor
 \If{$ind \neq \phi$}
 \State $S_{ind} \longleftarrow C_i^t$  \  \algocomment{//Replace checkpoint $ind\in\cS$ with new checkpoint}
  \State Update $\alpha$ ; Update $\xi$
 \EndIf \\
 return $\cS, \alpha, \xi$
 \end{algorithmic}
\end{algorithm}

\subsection{Data valuation using selected checkpoints}

In this section, we describe our algorithm for estimating the value for each training datapoint using the selected checkpoints and corresponding features selected by the CheckSel algorithm.
Given the set of indices $\cS$, the coefficients $\alpha_i$, and the corresponding feature vectors $C_i$, the final estimate of decrease in loss vector is given by:
\begin{align}
\label{eq:finalapp2}
\mathcal{I}_T = \sum_{i\in \cS} \alpha_i C_i
\end{align}

Let $\cB_j$ be the mini-batch associated with the selected checkpoint j $\in \cS$ and $z_j^i \in \cB_j$ is the i-th datapoint in $\cB_j$. 
Using equation \ref{eq:finalapp2}, we can assign the score or value to a training datapoint $z^i_j \in \cB$ as:
\begin{align}
\label{eq:scoreval}
Value(z_j^i \in \cB_j) = \frac{1}{|\cB_j|} \sum_{z' \in \cZ'} (\alpha_j \mathcal{C}_i^j(z'))
\end{align}
For datapoints $e \in \cZ \setminus \cB$, there is no direct information to estimate their value using the checkpoints. Hence for such datapoints, we find the nearest neighbouring point $z \in \cB$ (using feature similarity) for each element $e$ in $\cZ \setminus \cB$. We compute the value of $e$ as the value of $z$ using Equation \ref{eq:scoreval}. 
The intuition behind using nearest neighbour technique is that if the point $e$ (whose score is unknown) is distant from the nearest neighbour $z$ (among the set of known scored points) by a factor $\epsilon$, the change in total decrease in loss $I_T$ for including $e$ in place of $z$ is bounded. The following theorem states the bound.

\begin{theorem}
\label{th:score}
Let $e \in \cZ \setminus \cB$ be a training datapoint outside the minibatches corresponding to selected checkpoints, and $z$ is its nearest neighbor, such that dist($e$,$z$) $\le \epsilon$. Also, let $\theta_1,...,\theta_T$ be the training trajectory and the loss function of the neural network is $L$-smooth, i.e. its gradient is Lipschitz continuous for these parameter values.

Further, let $I_T(z') = \sum_{t=1}^T \eta_t \nabla_{\theta_t}l(Z).\nabla_{\theta_t}l(z')$ be the estimate of total decrease in loss for any test datapoint $z'$, and $I'_T(z')$ be the same when the training point $e$ is replaced by $z$. Then:
\begin{equation}
    | I'_T(z') - I_T(z') | \le O(T\epsilon)
\end{equation}
\end{theorem}

Proof of the theorem hinges on the fact that $I_T$ is a Lipschitz function of the gradients of loss function given the training trajectory, which in turn is a Lipschitz continuous function of the input data by assumption. In practise, we use the penultimate layer representation of the trained model to calculate the distance between training datapoints.


 \textbf{Simsel: diverse subset selection}: While the valuation scores derived above, or those given by TracIn evaluate the contribution of each training datapoint towards the decrease in loss, they do not automatically select the most diverse set of training datapoints for a given selection budget. Hence, we propose \textit{Simsel}, a data subset selection technique to incorporate diversity into the subset of size $f$. The key idea behind diverse selection is to calculate a vector of values of each training datapoint, corresponding to all validation datapoints. Then, similarity between two training datapoints in terms of their value can be defined as the cosine similarity between the value vectors, called \textit{contrib} vectors. Hence the contrib vector is defined as:
 \begin{align}
\label{eq:score}
contrib(z_j^i) = [ (\frac{1}{|\cB_j|} (\alpha_j \mathcal{C}_i^j(z'))) , \forall z'\in Z']
\end{align}
 
 We define an objective function for selecting the optimal subset in Eqn. \ref{eqn:optim} that follows a facility location paradigm aimed at minimising the dissimilarity between pairs of instances ($\mathcal{D}_{ij}$), and returns $f$ diverse representatives.

\begin{align}
\label{eqn:optim}
    \min_{p_{ij}} & \sum_{i,j=1}^{|Z'|} p_{ij} \mathcal{D}_{ij} \\
    \mbox{\textbf{sub.to}:} & p_{i,j} \in [0,1] \ \forall i,j\ ;\  \ \sum_{j=1}^{|Z'|} p_{i,j} = 1 \ \forall i \nonumber \\
    &\sum_{j=1}^{|Z'|}\| [p_{1,j} \dots p_{|Z'|,j} ] \|_2 \leq f \ \forall i
    \nonumber
\end{align}
We propose to solve the optimisation using Algorithm \ref{algo:simset} that selects instances in an online streaming setting by adapting a combination of addition and removal strategy.
\begin{algorithm}
 \caption{: Simsel}
 \label{algo:simset}
 \small
 \begin{algorithmic}
 \State \textbf{Input:}
 \State \hspace{2mm} Training dataset $\mathcal{Z}$ sorted based on obtained scores from Equation \ref{eq:scoreval}, Subset size $f$
 \State \textbf{Output:}
 \State \hspace{2mm} Subset of training data $Z_S \subseteq \cZ$
 \State{\textbf{Algorithm:}}\\
 \hspace{2mm} $Z_S^0 =$ Top $f$ elements from $\mathcal{Z}$ based on scores.\\
 \hspace{2mm} $\mathcal{S} = $   $\cos(contrib(z_1), contrib(z_2)) $ where $z_1,z_2 \in \cZ$\\
 \hspace{2mm} Let $B_1,...,B_L$ be the minibatches of $\cZ$. 
 \hspace{2mm}\For{each $B_l, l=1,...,L$ }\\
 \hspace{6mm}$Z' = Z_S^{l-1} \cup B_l$\\
 \hspace{6mm} Let $\mathcal{D}(z_1,z_2) =1 - \cS(z_1,z_2)$ for every pair $(z_1, z_2) \in Z'$ ; $\mathcal{D} \in \mathcal{R}^{|Z'| \times |Z'|}$\\
 \hspace{6mm} $Z_S^l = $ Top $f$ representatives from $Z'$ (Equation \ref{eqn:optim}).
 \EndFor
 \end{algorithmic}
\end{algorithm}

%% file: experiment1.tex
\section{Experimental Results}

We evaluate the effectiveness and efficiency of the proposed data valuation and subset selection methods in two settings: (1) \textit{Transductive}: where the data valuation is applied only to training datapoints which are used for training the end-task model, and (2) \textit{Domain Adaptation}: where the end-task model is trained on a source training dataset, while the data valuation is applied to a different but related target dataset.
For evaluation in the transductive setting (also followed by baseline methods \cite{NEURIPS2020_e6385d39}), we  use three standard datasets CIFAR10, CIFAR100, and Tinyimagenet \cite{Le2015TinyIV}, and a ResNet-18 model\cite{he2016deep} . Section \ref{sec:addgreater} demonstrates the overall effectiveness of the proposed methods,
and Section \ref{sec:analyse} analyses various properties of the selected data subsets to demonstrate their usefulness. 
Lastly, Section \ref{sec:domadapt} demonstrates the effectiveness of the proposed algorithms in the domain adaptation setting using the standard \textit{Office-Home} \cite{venkateswara2017deep} dataset. 


\subsection{Efficient data valuation}
\label{sec:addgreater}

In this section, we study the effectiveness of the proposed algorithms \textbf{CheckSel} (Algorithm \ref{algo:onlineomp}) and \textbf{SimSel} (Algorithm \ref{algo:simset}), and compare them with the  baseline methods \textbf{TracIn} \cite{NEURIPS2020_e6385d39} and \textbf{InfluenceFunc} \cite{koh2017understanding}. We report results with 3 methods / techniques derived from the proposed algorithms: (1) \textbf{CheckSel} - uses CheckSel for checkpoint selection, followed by the TracIn score for data selection, (2) \textbf{TracIn-SimSel} - uses uniformly selected checkpoints  like TracIn, followed by SimSel for data selection, and (3) \textbf{CheckSel-SimSel} - uses CheckSel for checkpoint selection followed by SimSel for data selection. 

\begin{table*}[]
\vspace{-5mm}
\caption{\textbf{Test set accuracy of models trained on selected set for various selection methods 
}}
\label{tab:sametime}
\centering
\begin{tabular}{|l||l|l||l|l|l|l|} \hline
 \textbf{Dataset}  & \textbf{\begin{tabular}[c]{@{}l@{}}Influence\\ Function\end{tabular}} & \textbf{TracIn} & \textbf{CheckSel} & \textbf{\begin{tabular}[c]{@{}l@{}}TracIn-\\ SimSel\end{tabular}} & \textbf{\begin{tabular}[c]{@{}l@{}}CheckSel-\\ SimSel\end{tabular}} \\ \hline
CIFAR10 (10\% \#CP = 10)  & 59.9 & 44.13  & \textbf{79.93} (\textbf{\textcolor{blue}{+35.8}})             & 78.63 (+34.5) & 79.6 (+0.97)  \\ \hline \hline
CIFAR100 (10\% \#CP = 10) & 21.82 & 21.48  & 35.48 (+14) & 33.96 (+12.48) & \textbf{35.76} (+1.8)                                                               \\ \hline \hline
Tinyimagenet (10\% \#CP = 10) & 11.74 & 20.62  & 28.98 (+8.36) & 23.46 (+2.84) & \textbf{29.18} (+5.72)                                                               \\ \hline
\end{tabular}
\end{table*}

Table \ref{tab:sametime} shows the test set accuracy for the proposed methods when the models are trained on the selected points. 
The size of the selected dataset are taken as 10\% of the total training data size, and we store 10 checkpoints during the training iterations. The parameters are chosen so that   all the methods  are executed  in  similar time ($\sim 3$ hours (CIFAR10 and CIFAR100) and $\sim 11.5$ hours (Tinyimagenet)). We see that the proposed methods: CheckSel, TracIn-SimSel and CheckSel-SimSel consistently perform better than the baseline methods across all the datasets. For CheckSel and CheckSel-SimSel the numbers in brackets denote percentage point improvements due to the CheckSel algorithm (i.e. over TracIn and TracIn-SimSel respectively), while for TracIn-SimSel, they denote the improvement due to the SimSel algorithm (i.e. over TracIn). We note that both CheckSel and SimSel individually show huge improvements ($\sim 30 \%$) over TracIn, while their combination shows milder gains over the individual methods (possibly due to reaching the limiting performance for the given size).


Table \ref{tab:perftime} shows a consolidated result for selecting a subset of 10\% data across three datasets. We perform a hyperparameter search over \#CP for each dataset and report their best accuracies along with time taken for each of the methods. The best performing accuracies are highlighted in bold. We can also observe that the second best performing accuracy is less than a margin of 1\%. We divide the total timing profile (\textbf{Total Time}) into three components: training time on whole dataset, checkpoint selection time during training (\textbf{CheckSel Time}) applicable for CheckSel and its variant, data valuation and subset selection time (\textbf{DV/SS Time}) applicable for all. Various sub-parts of these components will be holding valid for the different methods.

We show that the proposed method CheckSel or its variant CheckSel-SimSel surpasses the baselines Influence Function \cite{koh2017understanding} and TracIn \cite{NEURIPS2020_e6385d39} by a large margin in the classification task. The method TracIn-SimSel uses Algorithm \ref{algo:simset} to select instances in the data selection phase, post the method of uniformly selecting checkpoints thus performing better than TracIn\cite{NEURIPS2020_e6385d39}. An exceptional case occurs in CIFAR100 where TracIn-SimSel performs slightly better than CheckSel. However, it can be attributed to the SimSel component in the data valuation phase. Overall, we can see that the valuation methods CheckSel and CheckSel-SimSel mostly outperform all the other baselines across all the datasets.

\begin{table*}[]
\vspace{-5mm}
\centering
\caption{\textbf{Best Performance for DV/SS methods with Total Time, Checkpoint Selection Time, and DV Time.}}
\label{tab:perftime}
\begin{tabular}{|llllll|}
\hline
\multicolumn{1}{|l|}{\textbf{Metrics}}                                                    & \multicolumn{1}{l|}{\begin{tabular}[c]{@{}l@{}}\textbf{Influence}\\ \textbf{Function}\end{tabular}} & \multicolumn{1}{l|}{\textbf{TracIn}} & \multicolumn{1}{l||}{\textbf{CheckSel}} & \multicolumn{1}{l|}{\begin{tabular}[c]{@{}l@{}}\textbf{TracIn -}\\ \textbf{SimSel}\end{tabular}} & \begin{tabular}[c]{@{}l@{}}\textbf{CheckSel -}\\ \textbf{SimSel}\end{tabular} \\ \hline
\multicolumn{6}{|l|}{\textbf{Tinyimagenet}(10\%, \#CP = 20)}       \\ \hline
\multicolumn{1}{|l|}{Test-set Accuracy}      & \multicolumn{1}{l|}{11.74}                                               & \multicolumn{1}{l|}{22.84}  & \multicolumn{1}{l||}{29.44}    & \multicolumn{1}{l|}{23.56}                                                   & \textbf{29.52}                                                    \\ \hline
\multicolumn{1}{|l|}{Total time (hrs)}    & \multicolumn{1}{l|}{11.3}                                                & \multicolumn{1}{l|}{12.91}  & \multicolumn{1}{l||}{12.01}    & \multicolumn{1}{l|}{13.1}                                                    & 12.2                                                      \\ \hline
\multicolumn{1}{|l|}{Checkpoint Selection time (hrs)} & \multicolumn{1}{l|}{-}                                                   & \multicolumn{1}{l|}{-}      & \multicolumn{1}{l||}{1.5}      & \multicolumn{1}{l|}{-}                                                       & 1.5                                                       \\ \hline
\multicolumn{1}{|l|}{Data Valuation time (hrs)}    & \multicolumn{1}{l|}{2.7}                                                 & \multicolumn{1}{l|}{4.3}    & \multicolumn{1}{l||}{1.9}      & \multicolumn{1}{l|}{4.49}                                                    & 2.09                                                      \\ \hline
\multicolumn{6}{|l|}{\textbf{CIFAR10}(10\%, \#CP = 10)}    \\ \hline
\multicolumn{1}{|l|}{Test-set Accuracy}      & \multicolumn{1}{l|}{59.9}                                                & \multicolumn{1}{l|}{44.13}  & \multicolumn{1}{l||}{\textbf{79.93}}    & \multicolumn{1}{l|}{78.63}                                                   & 79.6                                                      \\ \hline
\multicolumn{1}{|l|}{Total time (hrs)}    & \multicolumn{1}{l|}{4.0}                                                 & \multicolumn{1}{l|}{3.13}   & \multicolumn{1}{l||}{3.33}     & \multicolumn{1}{l|}{3.29}                                                    & 3.49                                                      \\ \hline
\multicolumn{1}{|l|}{Checkpoint Selection time (hrs)} & \multicolumn{1}{l|}{-}                                                   & \multicolumn{1}{l|}{-}      & \multicolumn{1}{l||}{0.3}      & \multicolumn{1}{l|}{-}                                                       & 0.3                                                       \\ \hline
\multicolumn{1}{|l|}{Data Valuation time (hrs)}    & \multicolumn{1}{l|}{2.0}                                                 & \multicolumn{1}{l|}{1.0}    & \multicolumn{1}{l||}{0.9}      & \multicolumn{1}{l|}{1.16}                                                    & 1.06                                                      \\ \hline
\multicolumn{6}{|l|}{\textbf{CIFAR100}(10\%, \#CP = 5)}     \\ \hline
\multicolumn{1}{|l|}{Test-set Accuracy}      & \multicolumn{1}{l|}{21.82}                                               & \multicolumn{1}{l|}{22.04}  & \multicolumn{1}{l||}{35.28}    & \multicolumn{1}{l|}{35.44}                                                   & \textbf{36.6}                                                      \\ \hline
\multicolumn{1}{|l|}{Total time (hrs)}    & \multicolumn{1}{l|}{4.0}                                                 & \multicolumn{1}{l|}{2.84}   & \multicolumn{1}{l||}{3.6}      & \multicolumn{1}{l|}{3.01}                                                    & 3.76                                                      \\ \hline
\multicolumn{1}{|l|}{Checkpoint Selection time (hrs)} & \multicolumn{1}{l|}{-}                                                   & \multicolumn{1}{l|}{-}      & \multicolumn{1}{l||}{0.4}      & \multicolumn{1}{l|}{-}                                                       & 0.4                                                       \\ \hline
\multicolumn{1}{|l|}{Data Valuation time (hrs)}    & \multicolumn{1}{l|}{2.0}                                                 & \multicolumn{1}{l|}{0.54}   & \multicolumn{1}{l||}{0.9}      & \multicolumn{1}{l|}{0.71}                                                    & 1.06                                                      \\ \hline
\end{tabular}
\end{table*}

\begin{figure*}[h!]
    \centering
    \scriptsize
    \begin{subfigure}{0.5\columnwidth}
    \centering
    \includegraphics[width=\textwidth,height=3cm]{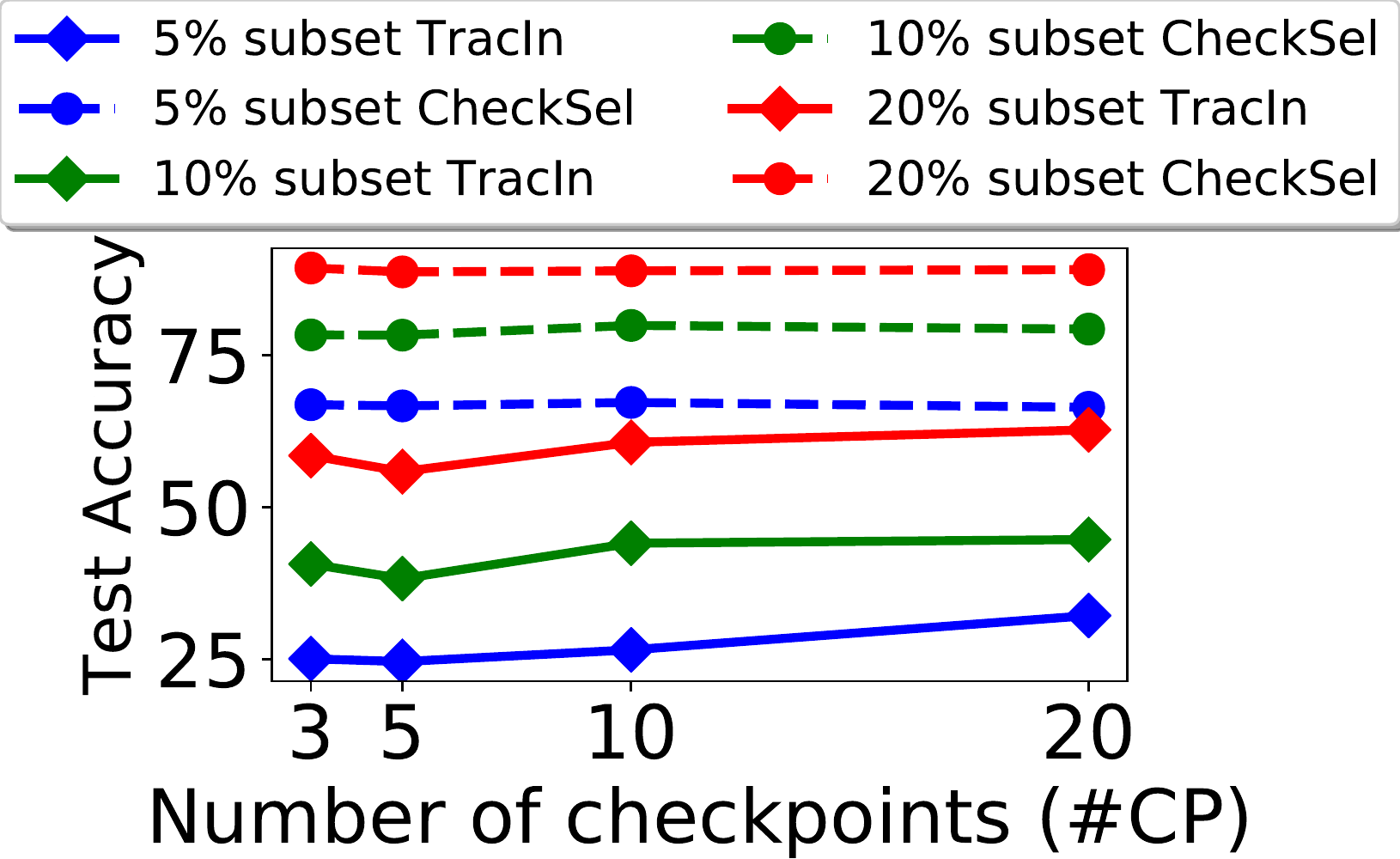}    
    \caption{}
    \end{subfigure}
    \begin{subfigure}{0.5\columnwidth}
    \centering
    \includegraphics[width=\textwidth,height=3cm]{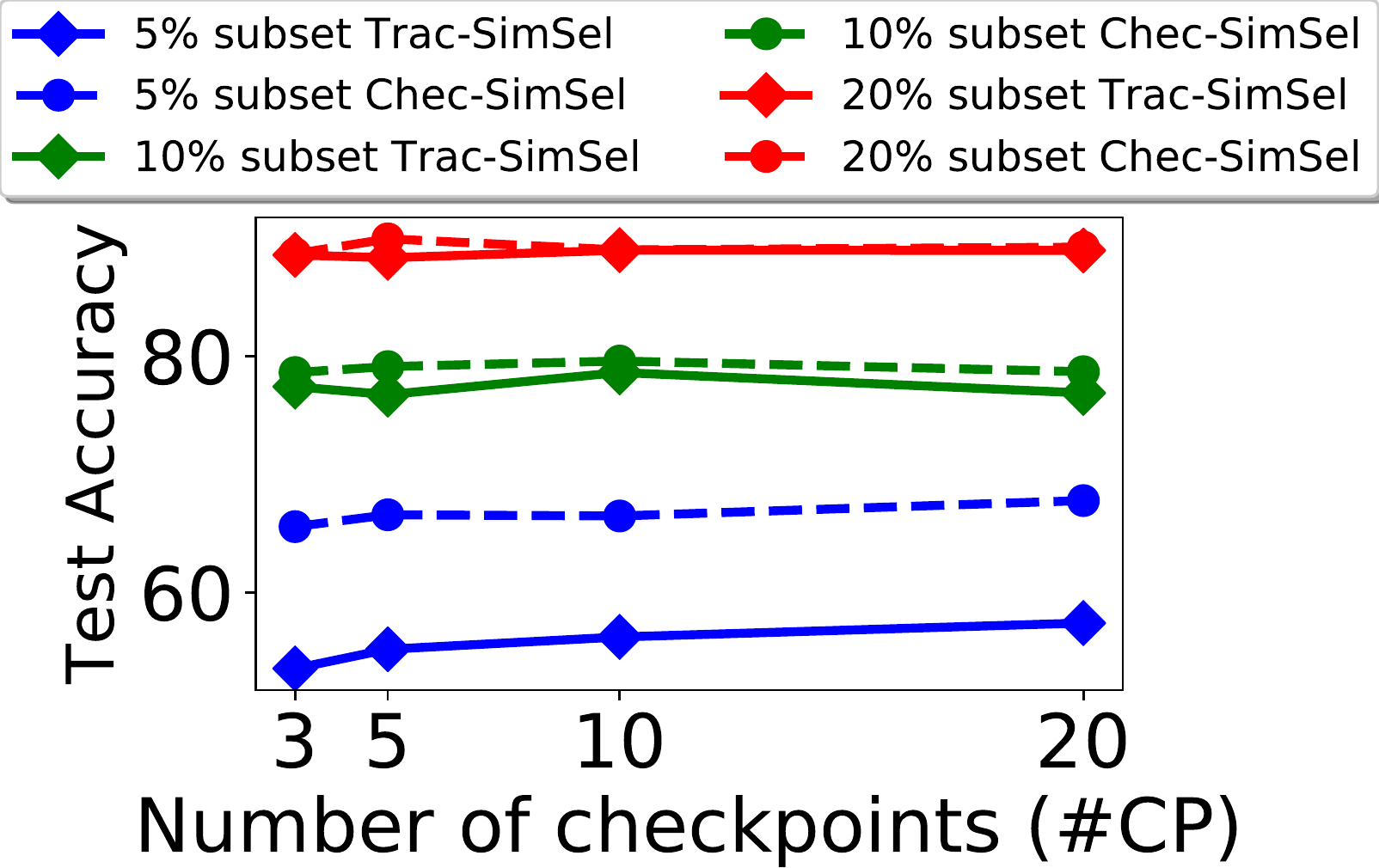}    
    \caption{}
    \end{subfigure}
    \begin{subfigure}{0.5\columnwidth}
    \centering
    \includegraphics[width=\textwidth,height=3cm]{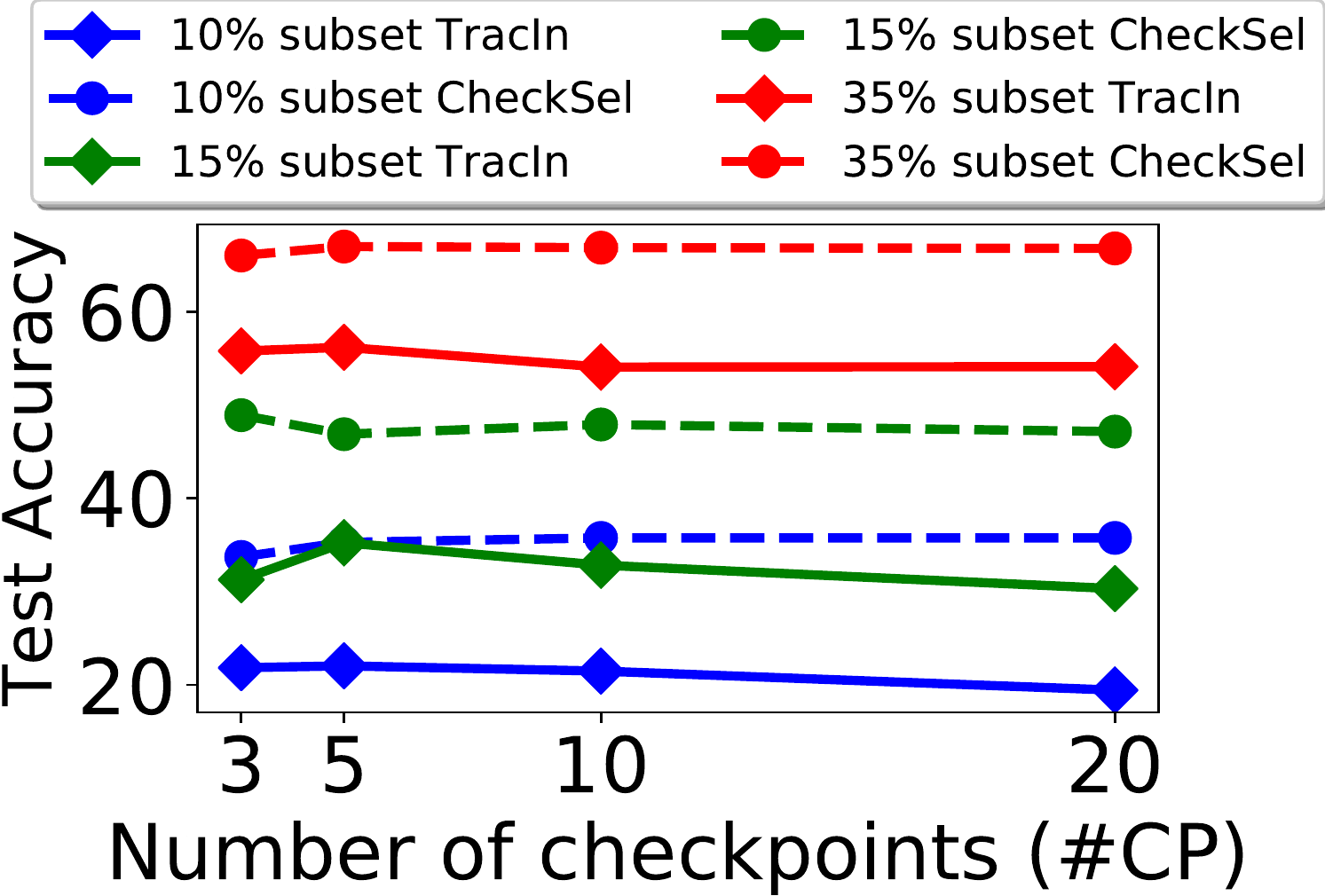}
    \caption{}
    \end{subfigure}
    \begin{subfigure}{0.5\columnwidth}
    \centering
    \includegraphics[width=\textwidth,height=3cm]{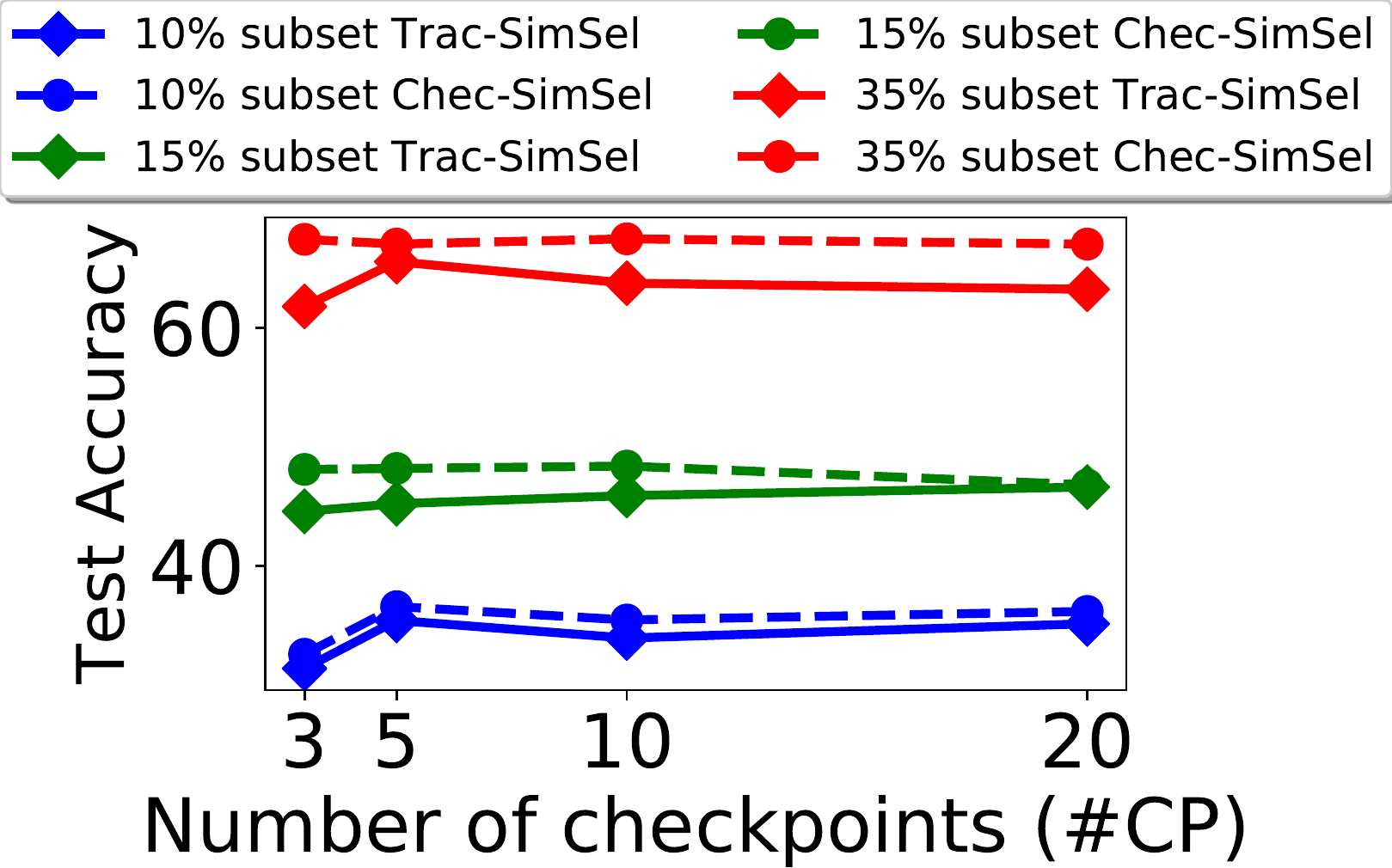}
    \caption{}
    \end{subfigure}
    \caption{\textbf{Accuracy comparison between TracIn \& CheckSel(a,c),TracIn-SimSel \& CheckSel-SimSel(b,d) for CIFAR10(a-b) and CIFAR100(c-d) respectively.}}
    \label{fig:varyperf}
\end{figure*}

We also show a variance across checkpoints (\#CP) and fractions of selected data in Figure \ref{fig:varyperf}. We compare the accuracies between (a) CheckSel and TracIn and (b) CheckSel-SimSel and TracIn-SimSel and can clearly observe that CheckSel and CheckSel-SimSel outperform the corresponding compared method. Alongside, it can be seen that the accuracy difference between the respective methods increases with decreasing fraction of subsets, thus denoting the accuracy of the subsets, even at a lower fraction.

\begin{figure*}[h!]
    \centering
    \begin{subfigure}{0.5\columnwidth}
    \centering
    \includegraphics[width=\textwidth,height=3cm]{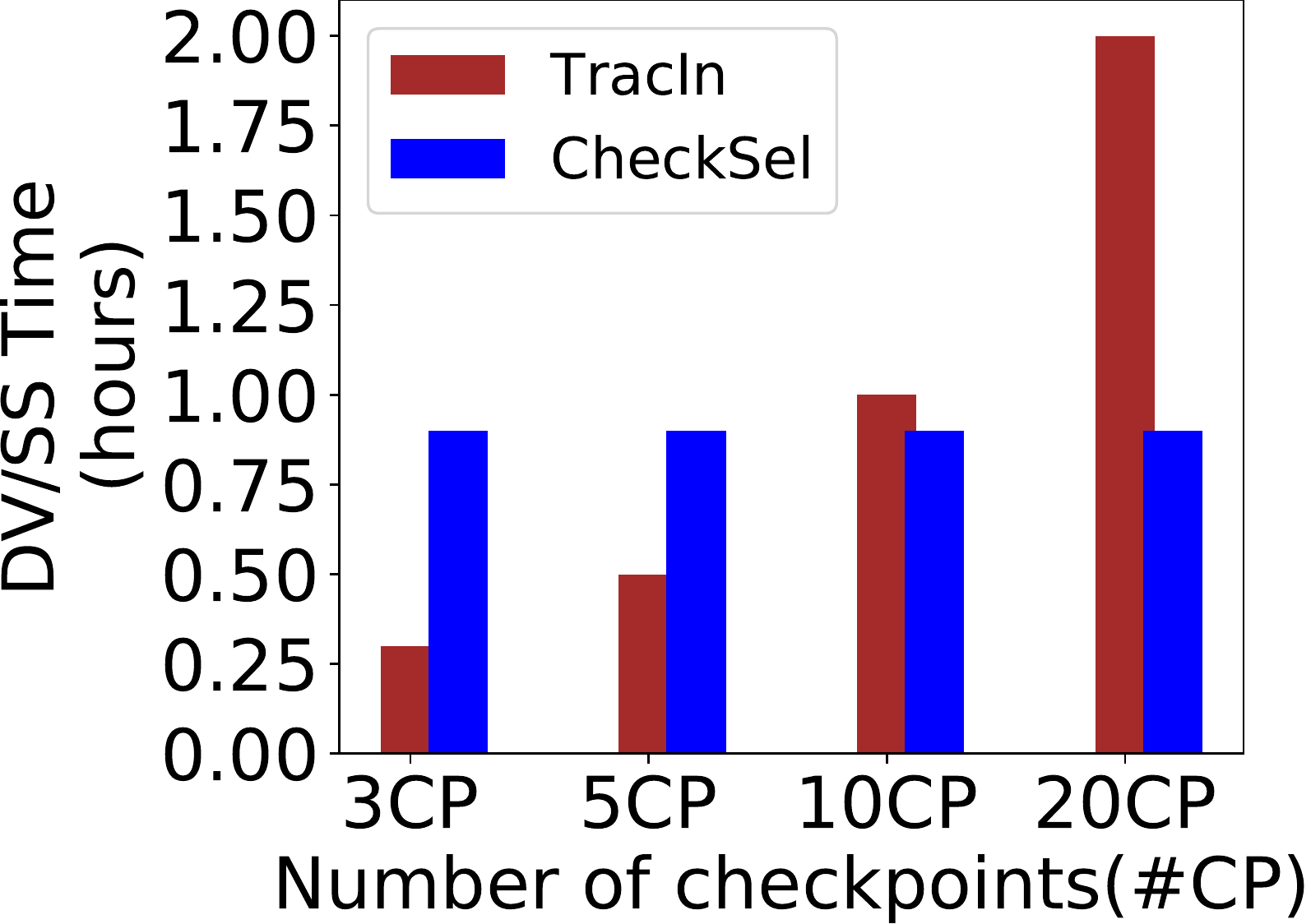}
    \end{subfigure}
    \begin{subfigure}{0.5\columnwidth}
    \centering
    \includegraphics[width=\textwidth,height=3cm]{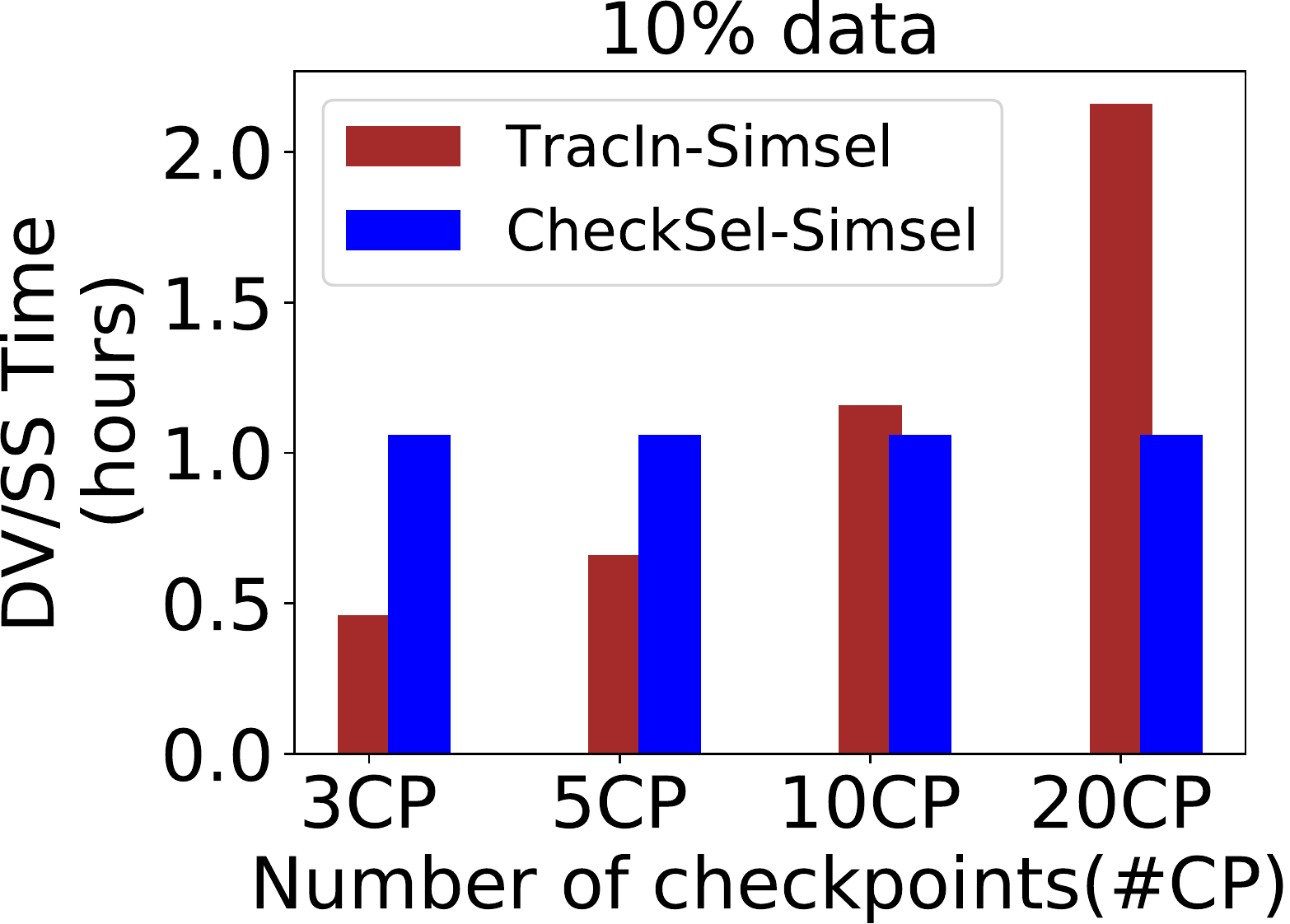}
    \end{subfigure}
    \begin{subfigure}{0.5\columnwidth}
    \centering
    \includegraphics[width=\textwidth,height=3.5cm]{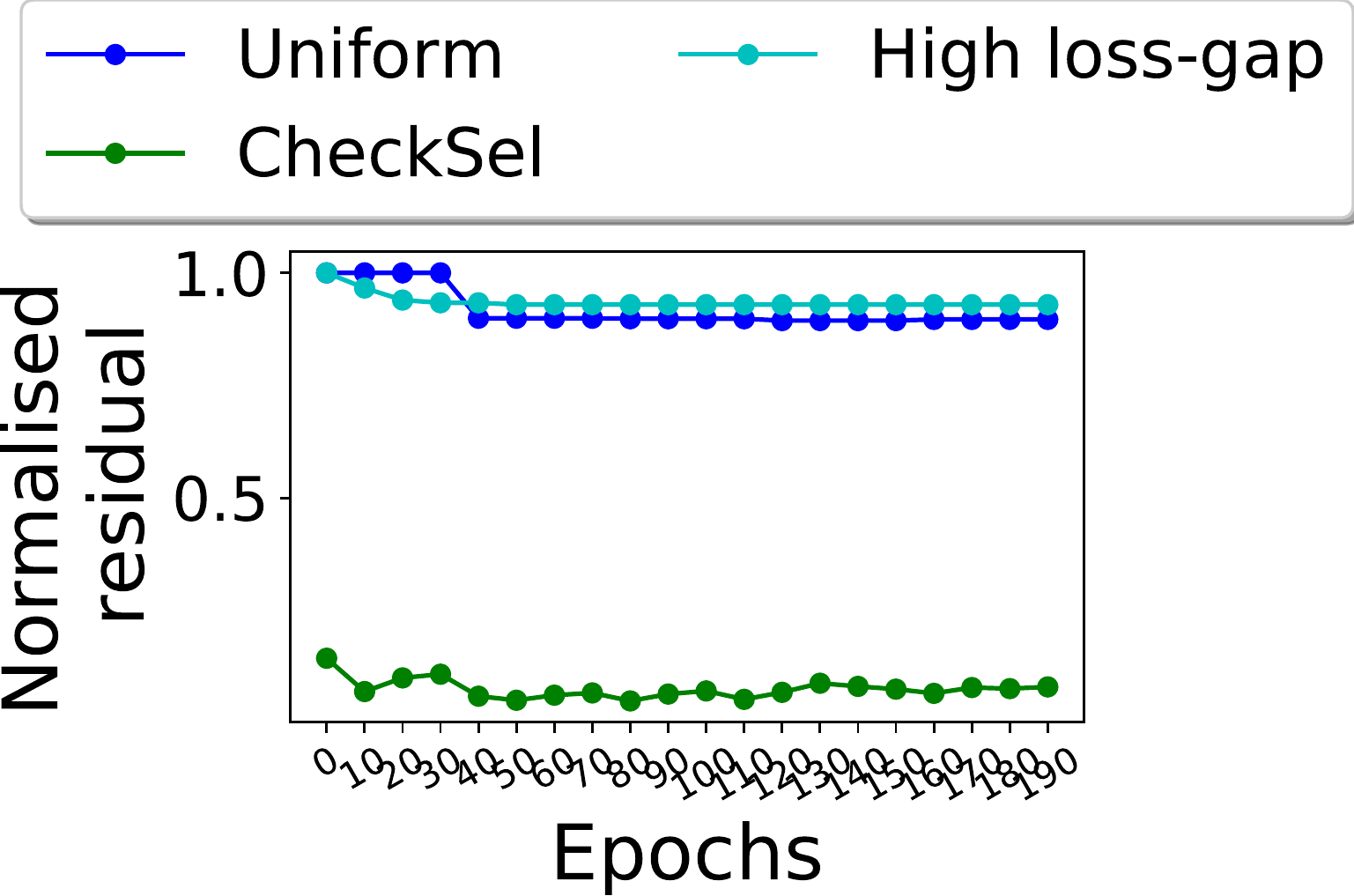}
    \end{subfigure}
    \caption{\textbf{CIFAR10: Comparison of DV/SS time across checkpoints between TracIn \& CheckSel(left) TracIn-SimSel \& CheckSel-SimSel(middle) for 10\% subset; (right)Normalised residual values for Uniformly selected, High loss-gap selected and CheckSel selected checkpoints.}}
    \label{fig:varytime1_residual}
    \vspace{-5mm}
\end{figure*}

While reporting the (DV/SS) time in Figure \ref{fig:varytime1_residual}(left,middle), we can see that with increasing value of checkpoints (\#CP), the time increases proportionally for TracIn and TracIn-SimSel. However, CheckSel and CheckSel-SimSel have similar value across all variations of checkpoints. This is because in CheckSel, the score of all training points are not computed using all the selected checkpoints but only with the checkpoint associated with that training datapoint. Hence, the valuation time (\textbf{DV/SS}) for CheckSel does not depend upon \#CP. The overhead component of checkpoint selection time (\textbf{CheckSel Time}) in variants of CheckSel is also not significantly high. We have added the timing profiles for the other two datasets in the Supplementary material. Overall, CheckSel or CheckSel-SimSel are as efficient as TracIn or TracIn-SimSel. Owing to the fact that our primary contribution lies in selecting checkpoints and CheckSel-SimSel performs as good as CheckSel, with SimSel incurring an additional cost of selection, we present the results ahead using TracIn and CheckSel.

We also show the normalised residual error (cumulative value function ($I_t$) - estimate) variation with epochs for uniform checkpoint selection, checkpoints with higher loss gap, and CheckSel based selection in Figure \ref{fig:varytime1_residual}(right). The normalised residual value for CheckSel is measured by $\frac{|I_{t} (z') - \sum_{i \in S^t}({\alpha_i S_i^t(z_i,z')})|_2}{|I_{t} (z')|_2}$ (notations follow from Section \ref{sec:method}) while for uniform and high loss gap selection, it is measured by $\frac{|I_{t} (z') - \eta \sum_{t'=1}^t \frac{1}{N} \sum_{i \in N}({\nabla l(\theta_{t'},z_i) \nabla l(\theta_{t'}, z')})|_2}{|I_{t} (z')|_2}$ where $t$ refer to the epoch under consideration. We observe that the difference between the residual values vary by a large gap, that gets manifested in the classification performance.

\subsection{Analysis of selected data}
\label{sec:analyse}

\begin{figure*}[h]
    \centering
    \begin{subfigure}{0.5\columnwidth}
    \centering
    \includegraphics[width=\textwidth,height=3cm]{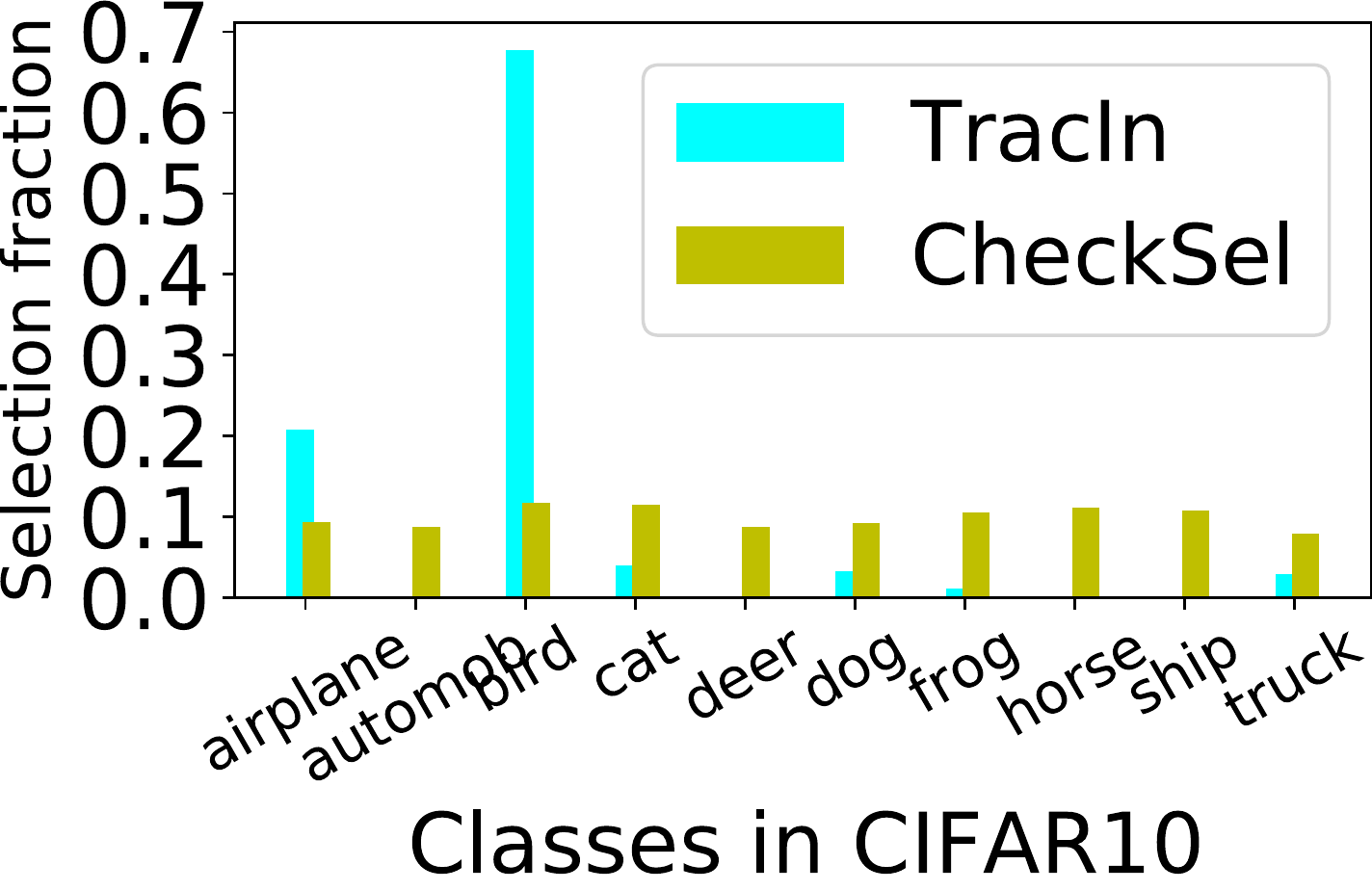}
    \end{subfigure}
    \begin{subfigure}{0.5\columnwidth}
    \centering
    \includegraphics[width=\textwidth,height=3cm]{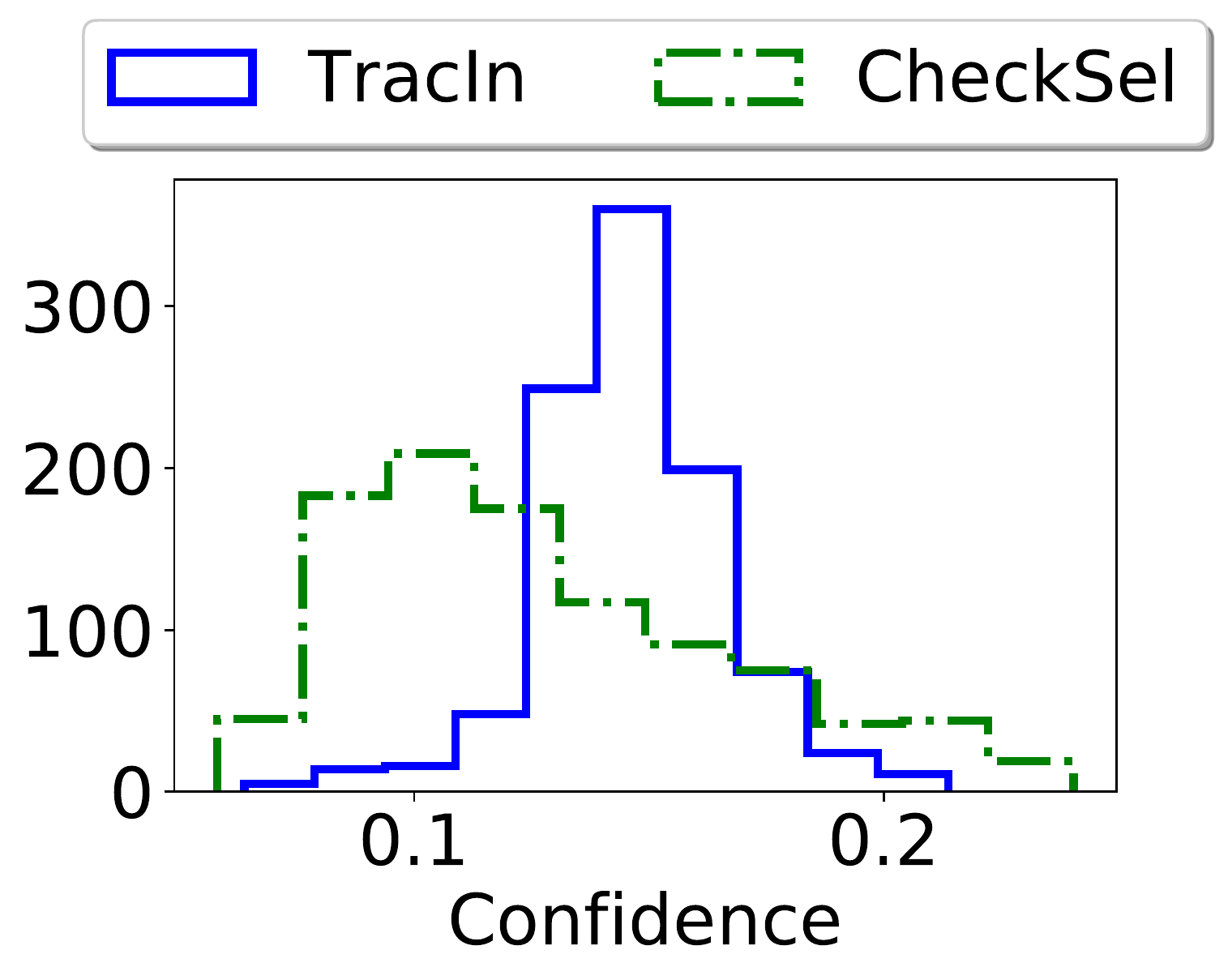}
    \end{subfigure}
    \begin{subfigure}{0.5\columnwidth}
    \centering
    \includegraphics[width=\textwidth,height=3cm]{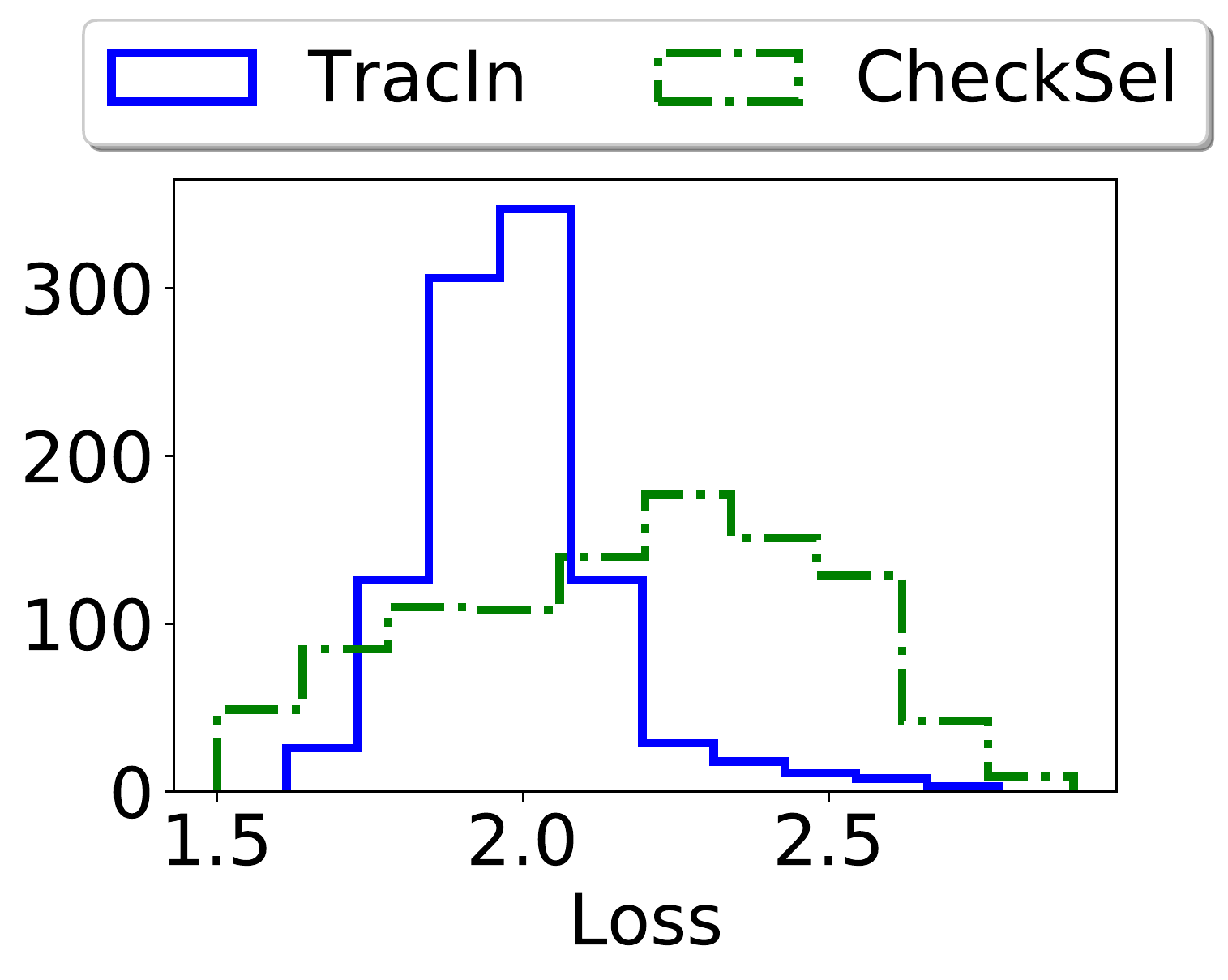}
    \end{subfigure}
    \caption{\textbf{(left)Fraction of top 1000 instances selected by TracIn and CheckSel ; Distribution of confidence values (middle) and loss values (right) of datapoints selected by TracIn and CheckSel in CIFAR10.}}
    \label{fig:subsetfrac_datamaps}
\end{figure*}

In this section, we intend to demonstrate the usefulness of the selected data through some data maps. We show the fraction of instances selected by TracIn \cite{NEURIPS2020_e6385d39} and CheckSel for CIFAR10 in Figure \ref{fig:subsetfrac_datamaps}(left), that clearly shows the diverse selection of CheckSel, across all classes leading to their better performance.

Inspired by \cite{swayamdipta2020dataset}, we construct data maps to visualise the selected dataset with respect to an underlying model. We use two measures for this purpose - \textit{Confidence} and \textit{Loss}. The intuition is that lower the confidence ($c$) or higher the loss ($l$) of a data point is, harder it is to be classified. We measure the confidence of a datapoint $x_i$ across $T$ epochs as:
\begin{equation}
    c_{x_i} = \frac{1}{T}\sum_{t=1}^T p_{\theta_t} (y_i|x_i)
\end{equation}
where $p_{\theta_t} (y_i|x_i)$ is the probability for $x_i$ to belong to the actual label $y_i$ at epoch $t$. In case of loss, we measure the usual cross-entropy loss for the point $x_i$. 

Selecting such harder examples (\textit{low confidence-high loss}) as a part of subset makes the model trained on subset, better thus leading to better performance. We can observe in Figure \ref{fig:subsetfrac_datamaps}(middle,right), the distribution of top 1000 datapoints selected by CheckSel from CIFAR10 have comparatively lower confidence and higher loss values, thus denoting the presence of harder examples. We can thus relate this back to their better performance accuracies. The results for CIFAR100 and Tinyimagenet are added to Supplementary.

\begin{figure*}[h!]
    \centering
    \begin{subfigure}{0.5\columnwidth}
    \centering
    \includegraphics[width=\textwidth,height=3cm]{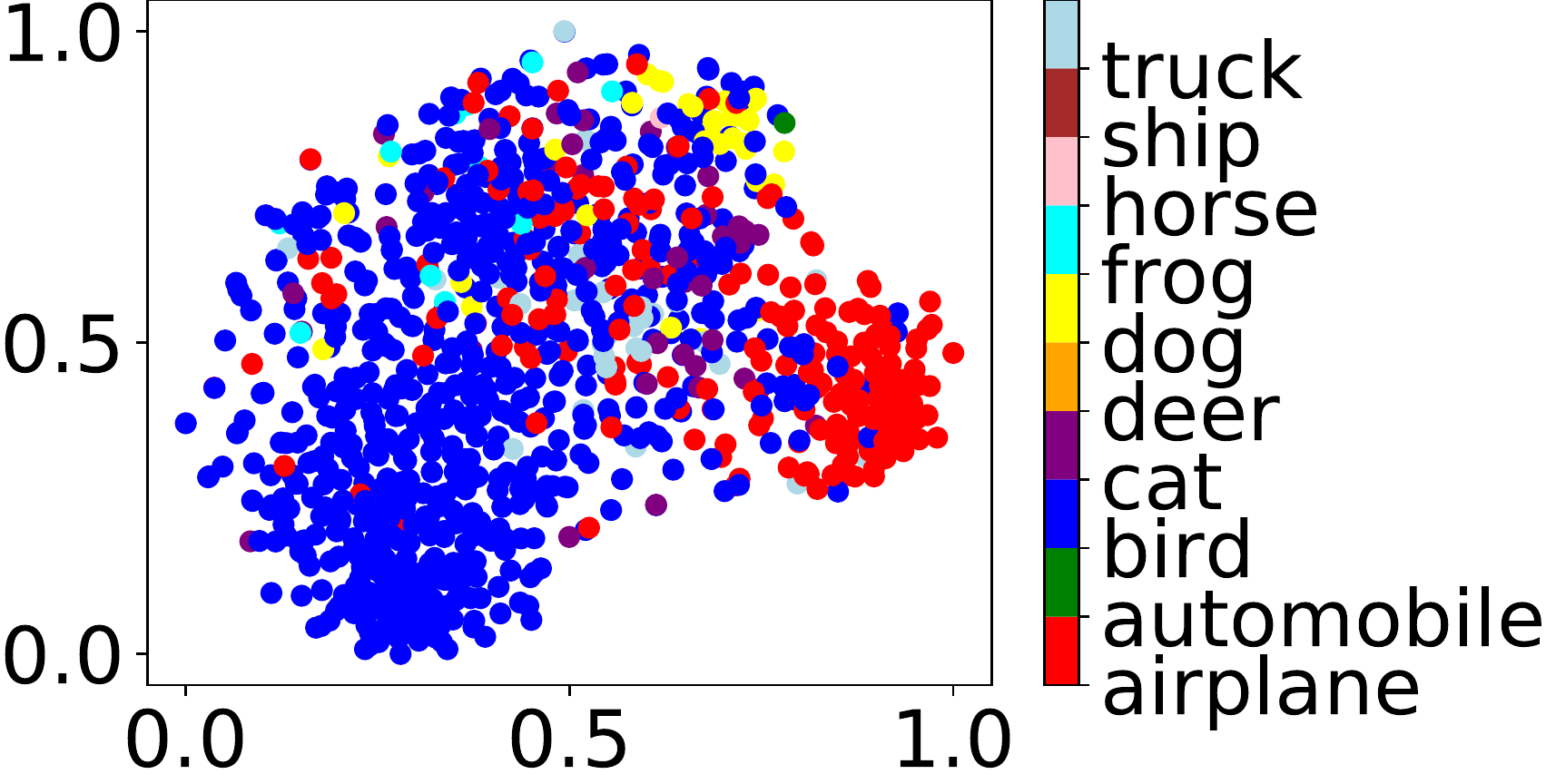}
    \end{subfigure}
    \begin{subfigure}{0.5\columnwidth}
    \centering
    \includegraphics[width=\textwidth,height=3cm]{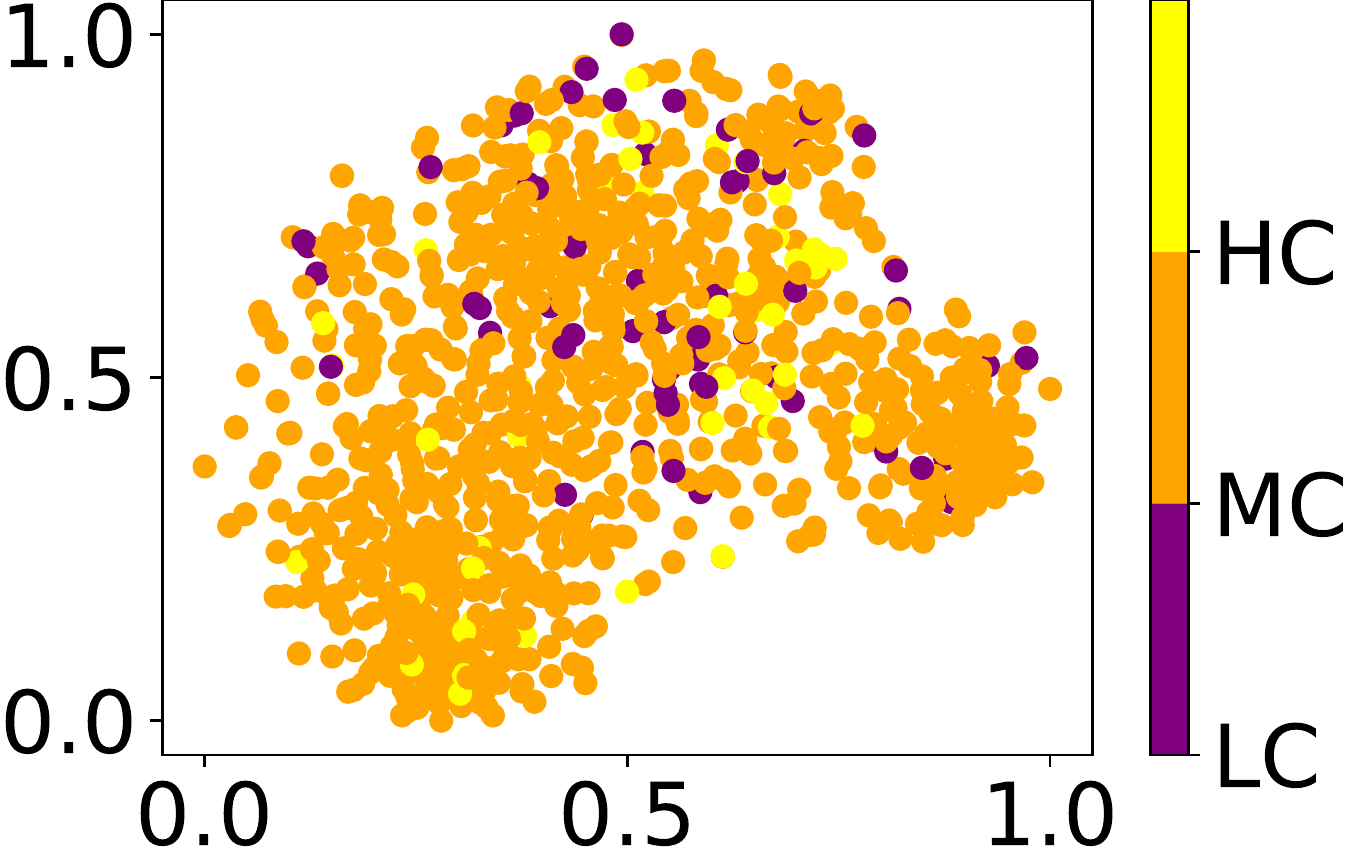}
    \end{subfigure}
    \begin{subfigure}{0.5\columnwidth}
    \centering
    \includegraphics[width=\textwidth,height=3cm]{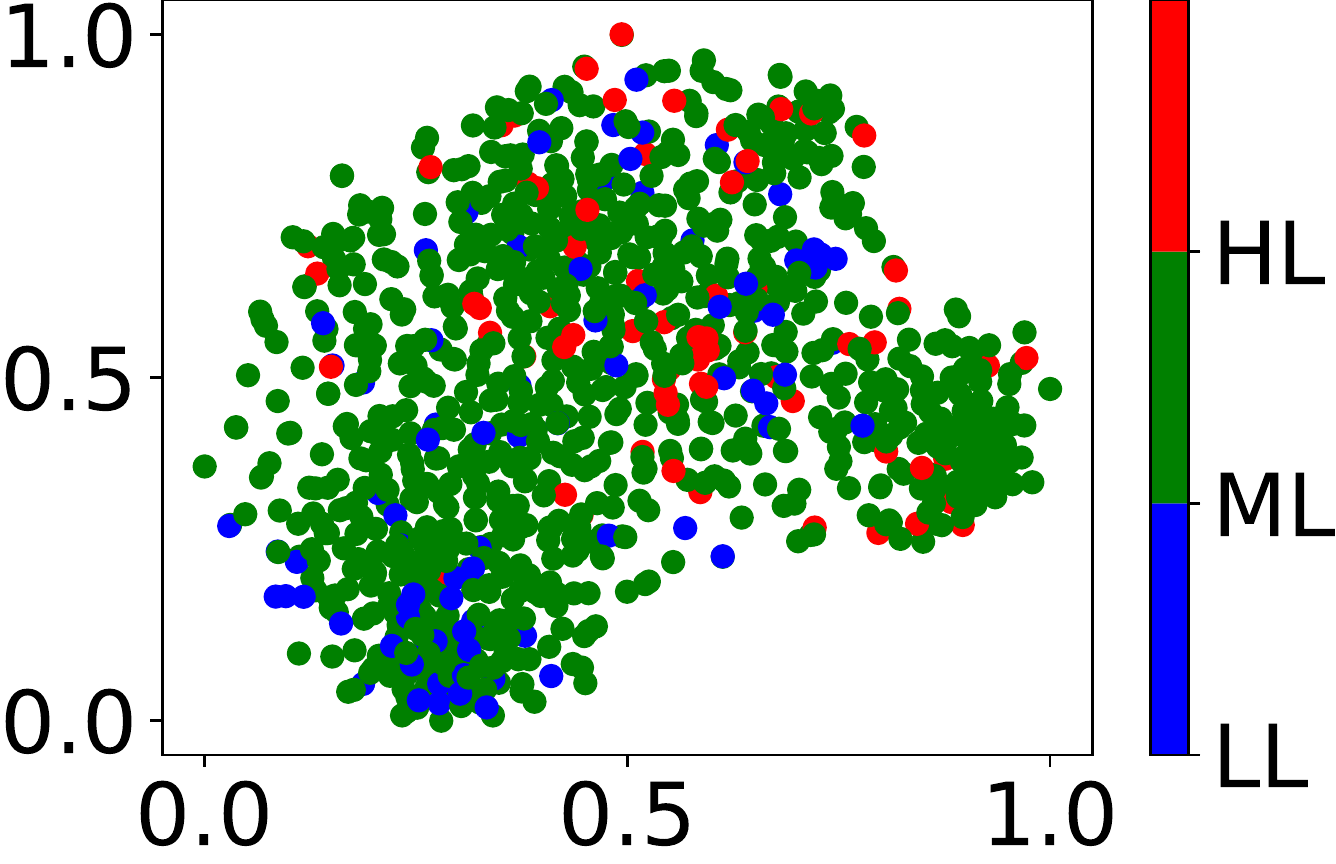}
    \end{subfigure}
    
    \begin{subfigure}{0.5\columnwidth}
    \centering
    \includegraphics[width=\textwidth,height=3cm]{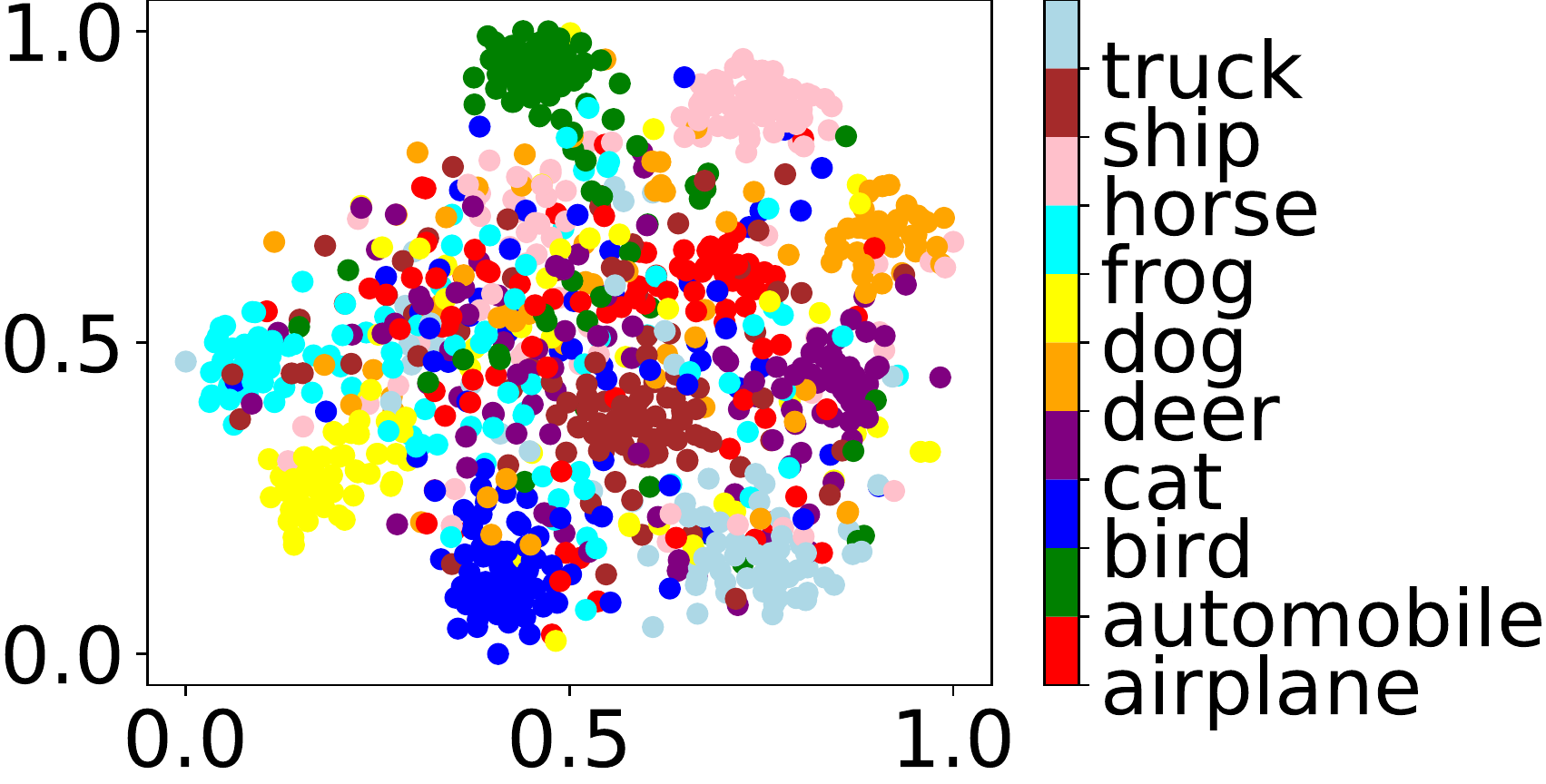}
    \caption{}
    \end{subfigure}
    \begin{subfigure}{0.5\columnwidth}
    \centering
    \includegraphics[width=\textwidth,height=3cm]{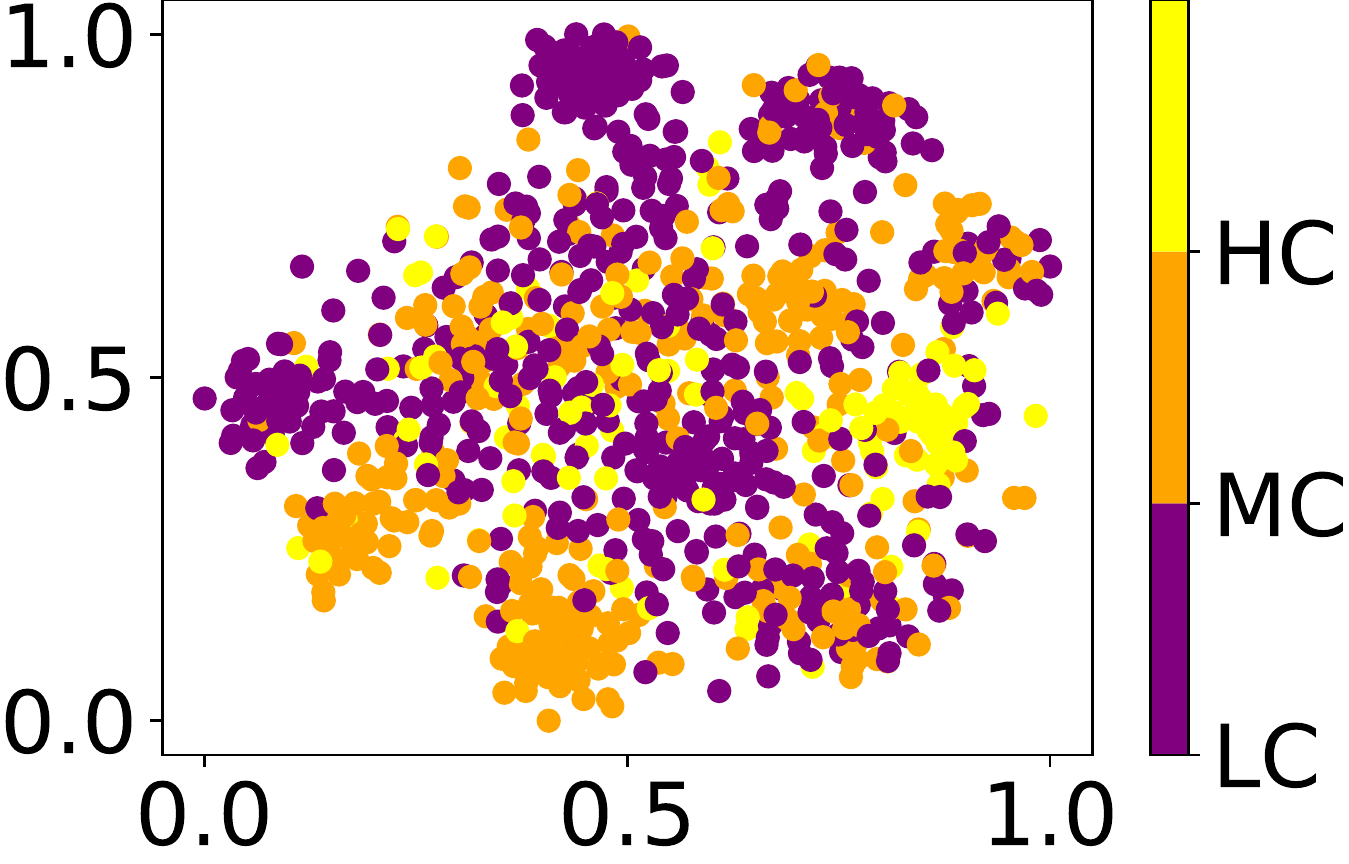}
    \caption{}
    \end{subfigure}
    \begin{subfigure}{0.5\columnwidth}
    \centering
    \includegraphics[width=\textwidth,height=3cm]{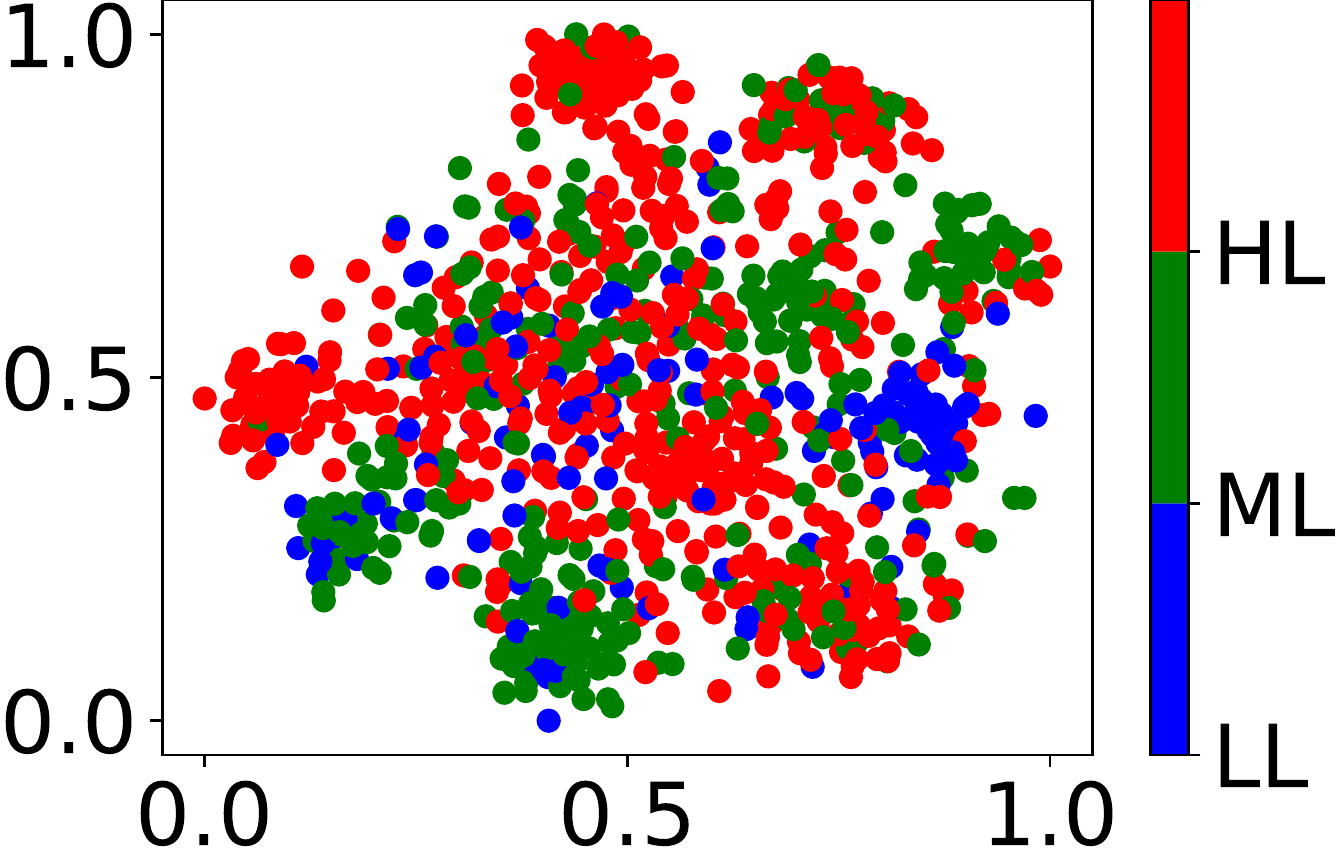}
    \caption{}
    \end{subfigure}
    \caption{\textbf{t-SNE of selected datapoints using TracIn (top) and CheckSel (bottom) based on classes(left),confidence(middle),and loss(right) values.}}
    \label{fig:subsetvis}
\end{figure*}

We show a t-SNE diagram in Figure \ref{fig:subsetvis} to visualise the top 1000 datapoints from the subsets obtained using TracIn and CheckSel from CIFAR10. We analyse the t-SNE from three dimensions - class, confidence, loss. We measure the confidence and loss values in 3 levels - High, Medium and Low. We can see that the left-side plots (TracIn) are having a majority of \textit{(bird)} instances (also seen in Figure \ref{fig:subsetfrac_datamaps}(left)), medium confidence (MC) and medium loss (ML) while the right-side plots (CheckSel) are having a fair mixture of all classes, majority of low confidence (LC) and high loss (HL) points, thus affirming the selection of valuable subsets. Another interesting observation for CheckSel based t-SNE plots are that instances from each of the classes are having a diverse level (\textit{high, medium, low}) of confidence and loss values.

\subsection{Data Valuation and Subset Selection in related Domains}
\label{sec:domadapt}

\begin{table*}[h]
\centering
\caption{\textbf{Validation Accuracy in Target domain for models  trained in various data valuation settings (20\% selection, 20 checkpoints)}}
\label{tab:daofficehome}
\begin{tabular}{|l||l|l||l|l|}
\hline
\textbf{Source-\textgreater{}Target}  & \textbf{TracIn} &\multicolumn{1}{|l||}{\textbf{CheckSel-Source}} & \multicolumn{1}{|l|}{\textbf{CheckSel-Target}} & \multicolumn{1}{|l|}{\textbf{CheckSel-Source-DA}} \\ \hline
Art-\textgreater{}ClipArt    & 31.4   & 47.99 (\textbf{+16.59}) & 54.97 & 53.07(-1.9)   \\ \hline
Art-\textgreater{}Product    & 33.95  & 70.11 (\textbf{+36.16}) & 75.23 & 71.28(-3.95) \\ \hline
Art-\textgreater{}Real World & 34.63  & 67.49 (\textbf{+32.86}) & 71.04 & 71.51(\textbf{+0.47}) \\ \hline
Art-\textgreater{}Art & 36.34   & 52.26(\textbf{+15.92}) & -  & - \\ \hline
\hline
ClipArt-\textgreater{}Art    & 36.34   & 51.61(\textbf{+15.27}) & 53.98 & 52.47(-1.51) \\ \hline
ClipArt-\textgreater{}Product    & 39.76  & 69.18(\textbf{+29.42}) & 75.74 & 69.81 (-5.93)\\ \hline
ClipArt-\textgreater{}Real World & 34.87  & 65.48(\textbf{+30.61}) & 74.05 & 65.48 (-8.57) \\ \hline 
ClipArt-\textgreater{}ClipArt & 33.09  & 52.3(\textbf{+19.21}) & -  & - \\ \hline \hline
Product-\textgreater{}Art    & 34.62   & 54.4(\textbf{+19.78}) & 56.55 & 57.42(\textbf{+0.87}) \\ \hline
Product-\textgreater{}ClipArt    & 30.85  & 57.32(\textbf{+26.47}) & 57.21 & 60.5(\textbf{+3.29})\\ \hline
Product-\textgreater{}Real World & 36.28  & 69.03(\textbf{+32.75}) & 73.04 & 71.16(-1.88) \\ \hline
Product-\textgreater{}Product & 35.81  & 65.59(\textbf{+29.78}) & -  & -\\ \hline  \hline
Real World-\textgreater{}Art    & 33.76   & 50.75(\textbf{+16.99}) & 56.56  & 55.48(-1.08) \\ \hline
Real World-\textgreater{}ClipArt    & 29.9  & 48.46(\textbf{18.56}) & 50.47 & 51.42(\textbf{+0.96}) \\ \hline
Real World-\textgreater{}Product & 33.25  & 64.41(\textbf{31.16}) & 70.97 & 68.6(-2.37)  \\ \hline
Real World-\textgreater{}Real World  & 35.22   & 60.4(\textbf{+25.18}) & -   & - \\ \hline
\end{tabular}
\end{table*}


Recall that the checkpoint selection algorithm, CheckSel, requires a valuation dataset which is used to calculate the \textit{value function}. In many practical applications, one is interested in data valuation in a related \textit{Target} domain dataset ($D_t$), even though the original checkpoints were selected using a value function from a \textit{Source} domain dataset ($D_s$). We call this the \textit{Domain Adaptation} setting for data valuation and data subset selection. In this section, we demonstrate the effectiveness of the proposed methods in the domain adaptation setting. The main advantage of this setting is that it avoids the relatively expensive training and checkpoint selection for the target domain.


For experimentation, we use the standard Office-Home dataset \cite{venkateswara2017deep} containing 4 domains - Art, Clipart, Product, and Real-World, each with 65 categories. We report results for all $16$ \textit{source -- target pairs} taken from the 4 domains. The experiments are conducted using a ResNet-50 model following standard procedure.
For every source -- target pair, both source dataset $D_s$ and   target dataset $D_t$ are divided into training and validation counterparts ($D_{st}$ , $D_{sv}$ and $D_{tt}$ , $D_{tv}$) respectively. 
In Table \ref{tab:daofficehome}, we report results for the following following methods:
\begin{enumerate}
\item \textbf{TracIn}: Checkpoints are sampled while training on $D_{st}$, Data-valuation is performed on $D_{tt}$ using TracIn score with valuation dataset $D_{tv}$.
\item \textbf{CheckSel-Source}: Checkpoints are selected using CheckSel, while training on $D_{st}$ using a value function given by $D_{sv}$. Data valuation is performed on $D_{tt}$ using TracIn score with valuation dataset $D_{tv}$. This represents the setting where target domain data is \textit{not} available during checkpoint selection, and no-retraining is needed for data valuation on target domain.
\item \textbf{CheckSel-Target}: Checkpoints are selected using CheckSel, while training on $D_{st}$ using a value function given by $D_{tv}$. Data valuation is performed on $D_{tt}$ using TracIn score with valuation dataset $D_{tv}$. This represents the unrealistic setting where target domain data is available during checkpoint selection, or the computationally expensive setting where re-training is performed on the full source dataset for data valuation on target domain.
\item \textbf{CheckSel-Source-DA}: Checkpoints are selected using CheckSel, while training on $D_{st}$ using a value function given by $D_{sv}$. Data valuation is performed on both $D_{st}$ and $D_{tt}$ using TracIn scores with valuation datasets $D_{sv}$ and $D_{tv}$, respectively. However, the final model is trained using selected subsets of both $D_{st}$ and $D_{tt}$ with the domain adaptation training technique described in \cite{Xu_2019_ICCV}. This setting is realistic since target domain data is not needed at the time of checkpoint selection, and it also avoids expensive re-training on the full-source dataset for target data valuation or subset selection.
\end{enumerate} 

Table \ref{tab:daofficehome} reports the validation accuracy on target domain using models trained on subsets of target domain training data and selected subset of source domain data (only for \textbf{CheckSel-Source-DA}).
We can observe that  \textbf{CheckSel-Source} performs substantially better than TracIn (numbers in brackets show change from TracIn), even on the target domain, hence confirming the robustness of the proposed CheckSel algorithm in the simple domain adaptation setting. Further, we see that \textbf{CheckSel-Target}, which has access to the target validation loss at the time of checkpoint selection performs significantly better than \textbf{CheckSel-Source}, hence demonstrating the utility of accurate value function estimate for checkpoint selection. However, this leads to a cumbersome requirement in real-life applications. The final proposed scheme \textbf{CheckSel-Source-DA} perform similarly to \textbf{CheckSel-Target} for most cases, and slightly worse in a few cases (the numbers in brackets show change from CheckSel-Target). Hence, we recommend using this scheme as it also retains the advantage of not having to re-train on full source dataset. In this setting, CheckSel takes 2x lesser running time for data valuation than TracIn for \#CP=20.

%% file: conclude.tex
\section{Conclusion}

In this paper, we propose a two-phase framework for efficient and accurate data valuation through checkpoint selection. We develop \textit{CheckSel}, an online OMP algorithm for selecting checkpoints, using which subsets of training data are found that best estimate the approximate decrease in validation loss. These subsets when trained from scratch outperform the baselines, sometimes by $\sim 30\%$ in terms of target accuracy. We also examine our algorithm on domain adaptation setting where we train on 20\% subset of target domain data that we find by selecting checkpoints during training on only source domain data and find it to outperform TracIn across all domains.

